%% file: main.tex
\documentclass{article} 
\usepackage{iclr2027_conference,times}

\input{math_commands.tex}

\usepackage{hyperref}
\hypersetup{colorlinks=true, urlcolor=blue}
\usepackage{url}
\usepackage{booktabs}
\usepackage{amsmath,amssymb}
\usepackage{graphicx}
\usepackage{xcolor}
\usepackage{enumitem}
\usepackage{float}
\usepackage[capitalise]{cleveref} 
\usepackage{subcaption} 

\newcommand{\drop}[1]{{\color{gray}\scriptsize(#1)}}
 
\title{UNREAL: Unifying Retrieval and Long-Context with a Single Model}

\author{Edan Kinderman$^1$,
Elad Hoffer$^1$,\,
Yochai Blau$^1$,\,
Brian Chmiel$^1$,\,
Ron Banner$^1$,\,
\\[0.15cm]
\textbf{Daniel Soudry$^{1,2}$,\,
Boris Ginsburg$^1$}
\\[0.15cm]
$^1$NVIDIA \quad $^2$Technion
\\[0.15cm]
\small{\texttt{\{{ekinderman},{ehoffer,}{yblau},{bchmiel},{rbanner},{dsoudry},{bginsburg}\}@nvidia.com}}
}

\iclrfinalcopy 
\begin{document}

\maketitle

\fancyhead{}
\renewcommand{\headrulewidth}{0pt}


\begin{abstract}
Long-context inference and Retrieval-Augmented Generation (RAG) handle evidence
selection at vastly different scales, from a single long prompt to an entire corpus. We ask
whether a single model-internal mechanism can select evidence across this
range. We introduce \textbf{UN}ifying \textbf{RE}trieval \textbf{A}nd \textbf{L}ong-Context with a Single Model (UNREAL), a model-native evidence selection framework to span corpus retrieval and long-context
inference. UNREAL encodes chunks and derives retrieval queries directly from
the frozen LLM's internal representations. It adds fewer than 500K
trainable parameters and leaves the backbone unchanged. On a 3B-token,
21M-chunk Wikipedia index, all four dense and hybrid UNREAL backbones outperform
state-of-the-art retriever-reranker systems. The best model
raises recall from 49.1\% to 73.2\% on
HotpotQA, from 31.7\% to 60.1\% on 2WikiMultiHopQA, and from 8.8\% to 14.4\%
on MuSiQue.
Applied to long-context tasks, the same selection mechanism removes distractors before generation, raising NoLiMa accuracy from 1.0\% to 24.83\% at its maximum context length of 128K tokens, and LV-Eval’s F1 score from 49.97\% to 54.66\% at 256K.
UNREAL also reduces FLOPs and time-to-first-token relative to full-context inference from roughly 32K tokens onward, with larger gains as context grows.
Together, these results establish model-internal evidence selection as a common foundation for corpus retrieval and evidence-sparse
long-context inference.

\end{abstract}

\begin{figure}[H]
  \centering
  \includegraphics[width=1.0 \linewidth ]{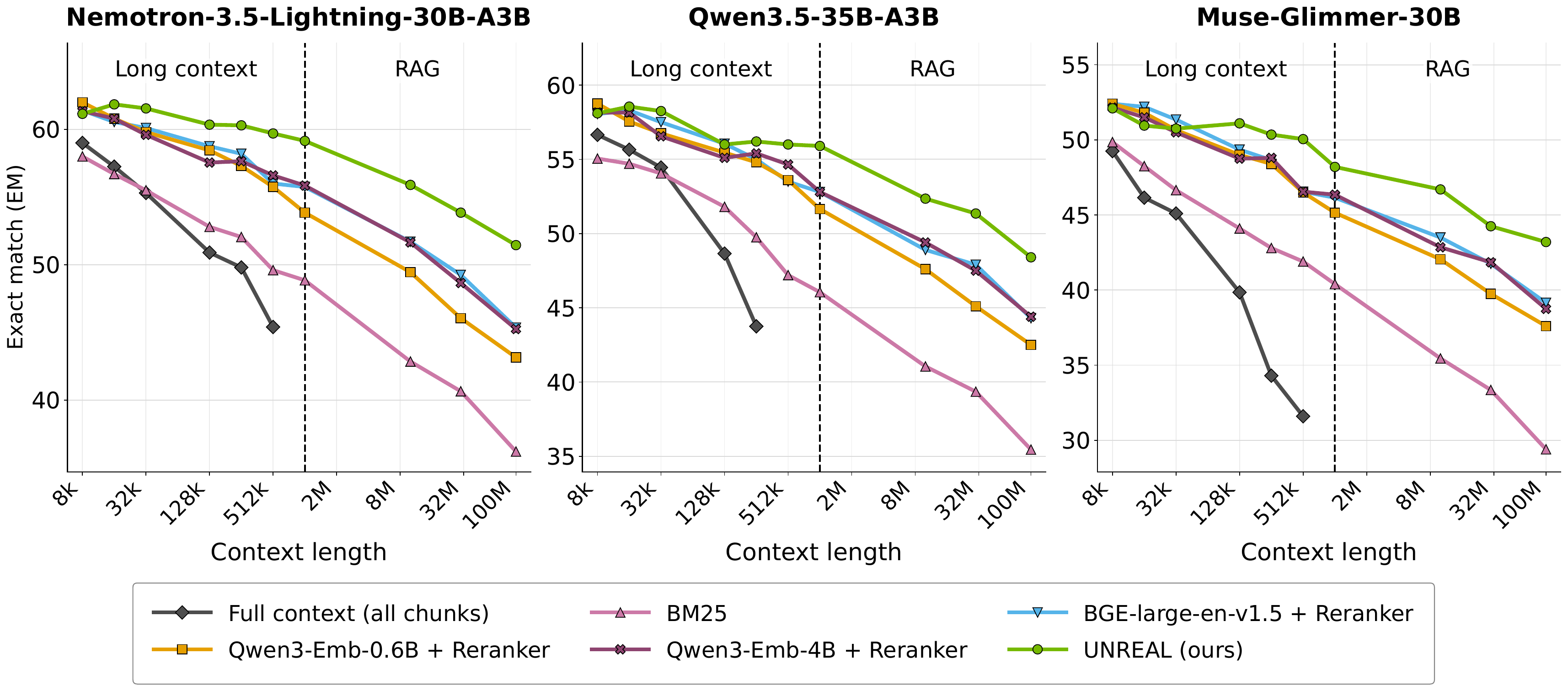}
\caption{
Exact match on question-answering as context length grows from 8K to 100M tokens.
Full context evaluation is shown only until computationally infeasible. For retrieval-based methods, relevant context is first extracted from the input and then fed to the model to generate the answer. Unlike baselines that rely on external models, UNREAL leverages the frozen LLM's internal representations to retrieve relevant context, maintaining the highest accuracy across long context lengths.}
  \label{fig:teaser-figure}
\end{figure}

\newpage

\section{Introduction}
\label{sec:intro}

\vspace{-.1cm}

Modern LLMs often process contexts where relevant information is sparsely distributed. To isolate these key signals, evidence selection is done differently, depending on the scale.
At long-context scale, selection is implicit and internal: the model receives the full context and must suppress irrelevant content while generating. This avoids having a separate retriever, but is vulnerable to distractors
\citep{modarressi2025nolimalongcontextevaluationliteral,shi2023distracted},
and scaling it requires dedicated architectures, efficiency mechanisms, and
long-context training.
At larger, corpus scale, this selection is explicit and external: a separate
retriever selects passages before generation
\citep{lewis2021retrievalaugmentedgenerationknowledgeintensivenlp}. This creates a
two-model pipeline that must be trained, served, and kept aligned.

Despite being studied as alternative approaches
\citep{xu2024retrievalmeetslongcontext,li2024selfroute,li2025lara,lee2024loft}, long-context and RAG perform the same operation: query-conditioned selection of candidate
evidence. Only the scale differs, from hundreds of chunks in a prompt to
millions in a corpus.
This suggests that the strengths of both approaches can be combined: make selection
internal, so evidence is ranked in the representation space of the model that
will use it, and make selection explicit, so distractors are removed during
generation \citep{yu2024defenseofrag}.
We therefore ask: can a single
pretrained LLM explicitly select evidence at both scales? An affirmative answer
would make retrieval and long-context inference two scales of one mechanism
rather than competing systems.

\subsection{Model-internal evidence selection across scales}
\label{sec:intro-retrieval}

\vspace{-.1cm}

We answer this question with \textbf{UN}ifying \textbf{RE}trieval \textbf{A}nd \textbf{L}ong-Context with a Single Model (UNREAL), a model-native evidence selector that operates across both long-context and corpus scales. We build on INTRA \citep{hoffer2026intra}, which demonstrated intrinsic retrieval
in encoder--decoder models. However, INTRA relies on a separate encoder and
cross-attention, which prevents its direct application to today's predominant
decoder-only LLMs. UNREAL removes this restriction by encoding candidate chunks
and deriving retrieval queries directly from the frozen LLM's internal states. UNREAL adds fewer than 500K trainable parameters, trained with contrastive loss, on top of the frozen LLM:
a soft prompt that elicits a retrieval mode and layer weights that combine its
internal states for evidence ranking.
Because it reads from the residual stream, the same mechanism applies across
dense-attention, linear-attention, and state-space architectures. 

At the corpus scale, UNREAL retrieves from a 3B-token, 21M-chunk Wikipedia index. All four tested backbones outperform the strongest dedicated retriever-reranker
systems. The best raises recall@10 from 49.1\% to 73.2\% on
HotpotQA and from 31.7\% to 60.1\% on 2WikiMultiHopQA.

The retrieval-trained UNREAL module can also rank chunks within a long-context prompt and return only the selected text to the LLM. This explicit removal of distractors raises accuracy on different long-context benchmarks: at the maximum evaluated length, NoLiMa accuracy rises from 1.0\% to 24.83\% at 128K tokens, while LV-Eval’s F1 score rises from 49.97\% to 54.66\% at 256K. This is achieved using the same LLM, without requiring an external retriever.
The method also reduces FLOPs and time-to-first-token versus full-context inference from roughly 32K tokens onward, with larger gains as context grows.
UNREAL thus turns corpus retrieval and long-context inference into two scales of the same model-internal selection mechanism (\cref{fig:teaser-figure}).

\subsection{Contributions}
\label{sec:intro-contributions}
\vspace{-.1cm}

\begin{enumerate}[leftmargin=1.5em]
  \setlength{\itemsep}{1pt}
  \item \textbf{Intrinsic corpus-scale retrieval with modern LLMs.} UNREAL extends INTRA
  \citep{hoffer2026intra} to decoder-only dense and hybrid LLMs, relying on a single
  frozen model to encode candidates and extract retrieval queries. With fewer than 500K trainable parameters, all four tested backbones outperform the strongest
  dedicated retriever--reranker systems on a 21M-chunk Wikipedia index
  (\cref{sec:retrieval}).
  \item \textbf{Long-context selection with the same mechanism.} 
  UNREAL outperforms both full-context inference and methods that rely on separate external retrievers across several long-context benchmarks, including NoLiMa, HELMET, LV-Eval, and LOFT, while reducing FLOPs and time-to-first-token at practical context lengths
  (\cref{sec:longctx}).
  \item \textbf{One selector across six orders of magnitude.} 
  The same model-native module selects evidence from an 8K-token context up to a 3B-token corpus, unifying long-context and retrieval reading without architectural changes or backbone finetuning.
\end{enumerate}

\begin{figure}[t]
  \centering
  \includegraphics[width=0.9 \linewidth ]{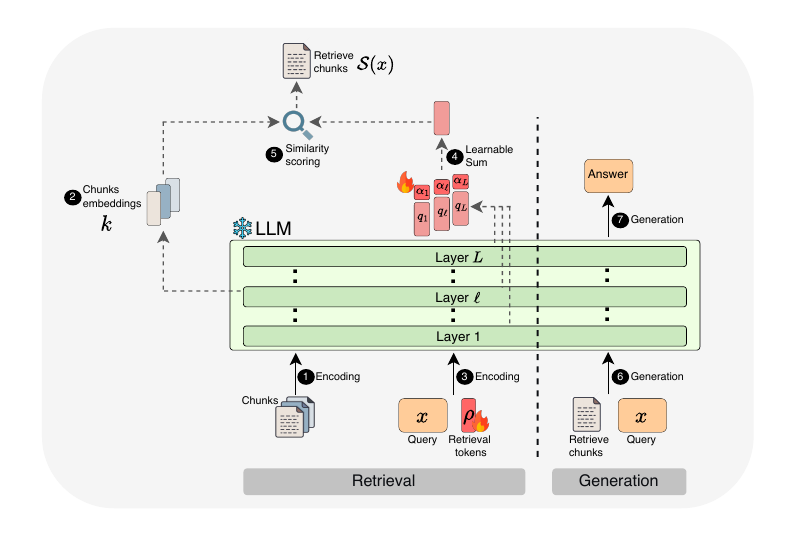}
   \vspace{-.6cm}
  \caption{UNREAL trains retrieval tokens $\rho$ and summation weights $\alpha$ (fire indicates training), while keeping the LLM frozen, enhancing its ability to identify relevant information in a context. Given a long context, these trained variables enable UNREAL to select relevant chunks and pass them to the same LLM for retrieval-augmented generation. Numbers indicate the order of operations.} 
  \vspace{-.3cm}

  \label{fig:intra-method-overview}
\end{figure}

\section{Method}
\label{sec:method}

Standard retrieval-augmented generation (RAG) \citep{lewis2021retrievalaugmentedgenerationknowledgeintensivenlp} delegates evidence selection to an external retriever, forcing production pipelines to maintain multiple models. In this work, we unify this process within a single decoder-only LLM using its intrinsic retrieval capabilities. The closest work to ours, INTRA \citep{hoffer2026intra}, asks whether an encoder--decoder model can instead retrieve
directly from its own encoded representations.
However, INTRA's reliance on an encoder--decoder architecture and cross-attention prevents its direct application to the more widely adopted decoder-only LLMs, limiting the range of pretrained models it can use.

In this work, we seek the same intrinsic retrieval capability in decoder-only LLMs, including hybrid architectures that interleave attention with other sequence mixers \citep{gu2024mamba}. 
Our method, UNREAL, generalizes INTRA by replacing its encoder-derived chunk
embeddings $k_i$ and cross-attention queries $q_\ell$ with representations
available inside a decoder-only model. UNREAL consists of three main components: chunk embeddings (\cref{sec:chunk-embeddings}), residual-state queries (\cref{sec:residual-state-queries}), and generation (\cref{sec:scoring-and-generation}). \cref{fig:intra-method-overview} provides an overview of the method.

\subsection{Chunk embeddings}
\label{sec:chunk-embeddings}

Let $\mathcal{C}=\{c_i\}_{i=1}^{M}$ denote a corpus of $M$ chunks. Given a query $x$, the goal is to retrieve the subset $\mathcal{S}(x)\subseteq\{1,\ldots,M\}$ containing the evidence needed to answer it. INTRA encodes each chunk once as $k_i=\operatorname{Enc}(c_i)$, where $\operatorname{Enc}$ denotes the model encoder. Decoder-only LLMs lack a separate encoder. UNREAL therefore uses the frozen LLM itself to encode each corpus chunk independently. For a chunk $c_i$ of length $T_i$, we select a single intermediate layer $\ell_c$ and extract its token-level representations:
\begin{equation} {k}_i \;=\; \operatorname{LLM}_{\ell_c}(c_i) \;\in\; \mathbb{R}^{T_i \times d}, \label{eq:method-chunk-rep} \end{equation} where $\operatorname{LLM}_{\ell_c}(\cdot)$ denotes the residual-stream states at layer $\ell_c$, and $d$ is the model's hidden dimension. Thus, each chunk is represented by a sequence of $T_i$ token vectors. 

Prior work has shown that intermediate layers often encode richer semantic representations than final layers \citep{Skean2025LayerBL}. Motivated by this observation, we evaluate representations from different layers in our corpus-retrieval setting and select $\ell_c$ according to performance on a development set (see the ablation in \cref{tab:corpus_layer_ablation}).

\subsection{Residual state queries}
\label{sec:residual-state-queries}

In order to perform query conditioned matching, UNREAL concatenates $R$ learned retrieval tokens $\{\rho_i \in \mathbb{R}^d\}_{i=1}^R$ to the query token embeddings $\{x_t \in \mathbb{R}^d\}_{t=1}^{T_q}$, where $T_q$ is the query length.
It also augments the input with an initial context $C_0(x)$, which represents the chunks ranked highest for the query $x$ by BM25 \citep{robertson2009bm25}.
Together, we get
\begin{equation}\label{eq:x-rho}
  x_{\mathrm{ret}}
  =
  \bigl[C_0(x),x_1,\dots,x_{T_q},\rho_1,\dots,\rho_R\bigr],
\end{equation}
Then it reads the residual stream states at the retrieval token positions:
\begin{equation}
  \forall \ell: \quad {q}_\ell(x_{\mathrm{ret}})
  \;=\;
  \big[ \operatorname{LLM}_\ell(x_{\mathrm{ret}})
  \big]\big|_{\left(|x_{\mathrm{ret}}|-R+1\right):|x_{\mathrm{ret}}|}
  \;\in\; \mathbb{R}^{R \times d}.
  \label{eq:method-query-rep}
\end{equation}
Placing the retrieval tokens last ensures that, under the causal mask, their states condition on both the query and the initial context. These internal layer-wise query representations ${q}_\ell$ are used to score every chunk $c_i$ with the late-interaction MaxSim operator \citep{colbert}:
\begin{equation}\label{eq:si}
  s_i(x;\rho,\alpha)
  \;=\;
  \operatorname{MaxSim} \Big( \sum_\ell \alpha_\ell {q}_\ell(x_{\mathrm{ret}}), {k}_i \Big),
\end{equation}
where $\alpha_\ell$ are learned layer-mixing coefficients and
$\operatorname{MaxSim}(u,v)\triangleq
\sum_a\max_b\langle u_a,v_b\rangle$.

The retrieved chunk set is then selected by
\begin{equation}
    \mathcal{S}_{\mathrm{UNREAL}}(x)
    =
    \left\{ i \in \{1,\ldots,M\} : s_i(x) \text{ is among the top-}n \text{ scores} \right\}.
  \label{eq:intra-background-score}
\end{equation}
Retrieval is therefore based on the internal representations ${q}_\ell$ and internal chunk encodings ${k}_i$ instead of an external index and model. Only $\rho$ and $\alpha$ are added as learned parameters, the LLM remains frozen.

Because our read-out operates on the residual stream, it can be applied uniformly across architectures that use softmax attention, linear attention, or state-space layers, including hybrids that combine these layer types. Aggregating read-outs across these layer types is
further motivated by the causal attention interpretation of selective SSMs
\citep{ali2025hiddenattention,jiang2026statefultokenreduction}.

\textbf{Retrieval training.}
The retrieval loss is a multi-positive InfoNCE objective
\citep{oord2018representation, chen2020simple} that contrasts oracle chunks
with sampled alternatives. For a query $x$, let $\mathcal{O}(x)$ contain the
oracle indices, corresponding to chunks that contain the annotated evidence
needed to answer the query. Let $\mathcal{N}(x)$ contain the hard negatives chunks. Over $\mathcal{B}(x)=\mathcal{O}(x)\cup\mathcal{N}(x)$, we minimize 
\begin{equation}
  \mathcal{L}_{\mathrm{retrieval}}
  =
  -\frac{1}{|\mathcal{O}(x)|}
  \sum_{j\in\mathcal{O}(x)}
  \log
  \frac{\exp({s}_j(x)/\tau)}
       {\sum_{i\in\mathcal{B}(x)}\exp({s}_i(x)/\tau)},
\end{equation}
where $\tau$ is the temperature. Training updates only the retrieval tokens
$\rho_i$ and layer-mixing coefficients $\alpha_\ell$, while the LLM parameters remain frozen. We then apply the resulting selector to retrieval tasks in \cref{sec:retrieval} and long-context benchmarks in \cref{sec:longctx}.

\textbf{Compression.} To reduce corpus-scale storage, we partition the full
token-level chunk representations $k_i\in\mathbb{R}^{T_i\times d}$ into
$L_p$ contiguous groups and mean-pool each group, compressing chunks from
$T_i\times d$ to $L_p\times d$. Similarly, we mean-pool the $R$
retrieval-token states $q_\ell$ across $G$ groups, compressing queries
from $R\times d$ to $G\times d$. The parameters $L_p$ and $G$ trade off
retrieval accuracy against index size and MaxSim compute cost.

\subsection{Scoring and generation}
\label{sec:scoring-and-generation}

Following
\cref{eq:intra-background-score},
$\mathcal{S}_{\mathrm{UNREAL}}(x)$ contains the indices of the $n$
highest-scoring chunks. To provide the model with this relevant context, the selected chunks
$C_{\mathrm{UNREAL}}(x)=[c_i : i\in\mathcal{S}_{\mathrm{UNREAL}}(x)]$ are
concatenated with the query in the LLM prompt:
\begin{equation}
  y
  \;=\;
  \operatorname{LLM}\!\left(
    [\,C_{\mathrm{UNREAL}}(x)\,,\,x\,]
  \right).
  \label{eq:intrad-generation}
\end{equation}
Unlike INTRA, which passes the selected encoder memories through
cross-attention, UNREAL re-encodes the
selected chunks during generation. Inference thus uses one LLM forward pass to
form the retrieval queries ${q}_\ell$, and a second ordinary generation pass over the
selected chunks.

In the following sections, we present how UNREAL can be used for both corpus retrieval and long-context evidence selection. We then discuss the connection between retrieval and long-context tasks.

\section{UNREAL as a Retrieval Mechanism}
\label{sec:retrieval}
\vspace{-.3cm}

Our experiments evaluate whether UNREAL remains effective at a scale requiring genuine retrieval over the 3B-token, 21M-chunk Wiki-2018 corpus \citep{karpukhin-etal-2020-dense}, with a mean chunk length of 140 tokens. 
We report complete-evidence recall@$n$: the fraction of examples with \emph{all} annotated oracle chunks retrieved in the top-$n$ results. This is particularly demanding for multi-hop tasks, where partial evidence receives no credit. Average recall@$n$ appears in \cref{fig:intra-avg-recall-10,fig:intra-avg-recall-20}.

\textbf{Benchmarks and baselines.}
The evaluation spans HotpotQA, 2WikiMultiHopQA, MuSiQue, HoVer, IIRC, SQuAD
v2, FEVER, and Natural Questions
\citep{yang2018hotpotqa,ho2020constructing,trivedi2022musique,jiang2020hover,ferguson2020iirc,rajpurkar2018squadv2,thorne2018fever,kwiatkowski2019natural}.
We map oracle chunks to the shared Wiki-2018 corpus using a KILT-style
procedure \citep{petroni2021kilt}.
We compare against BM25, Qwen3-Embedding and BGE dense retrievers, the Jina
cross-encoder reranker, LightOn multi-vector (late-interaction) retriever,
and hybrid RAG based on RRF
\citep{robertson2009bm25,zhang2025qwen3embedding,xiao2024cpack,jina2024rerankerv2,sourty2026denseonlateon,cormack2009rrf}.
See Appendix~\ref{app:experimental-setting} for the full details.

\begin{figure}[t]
  \centering
  \includegraphics[width=\linewidth]{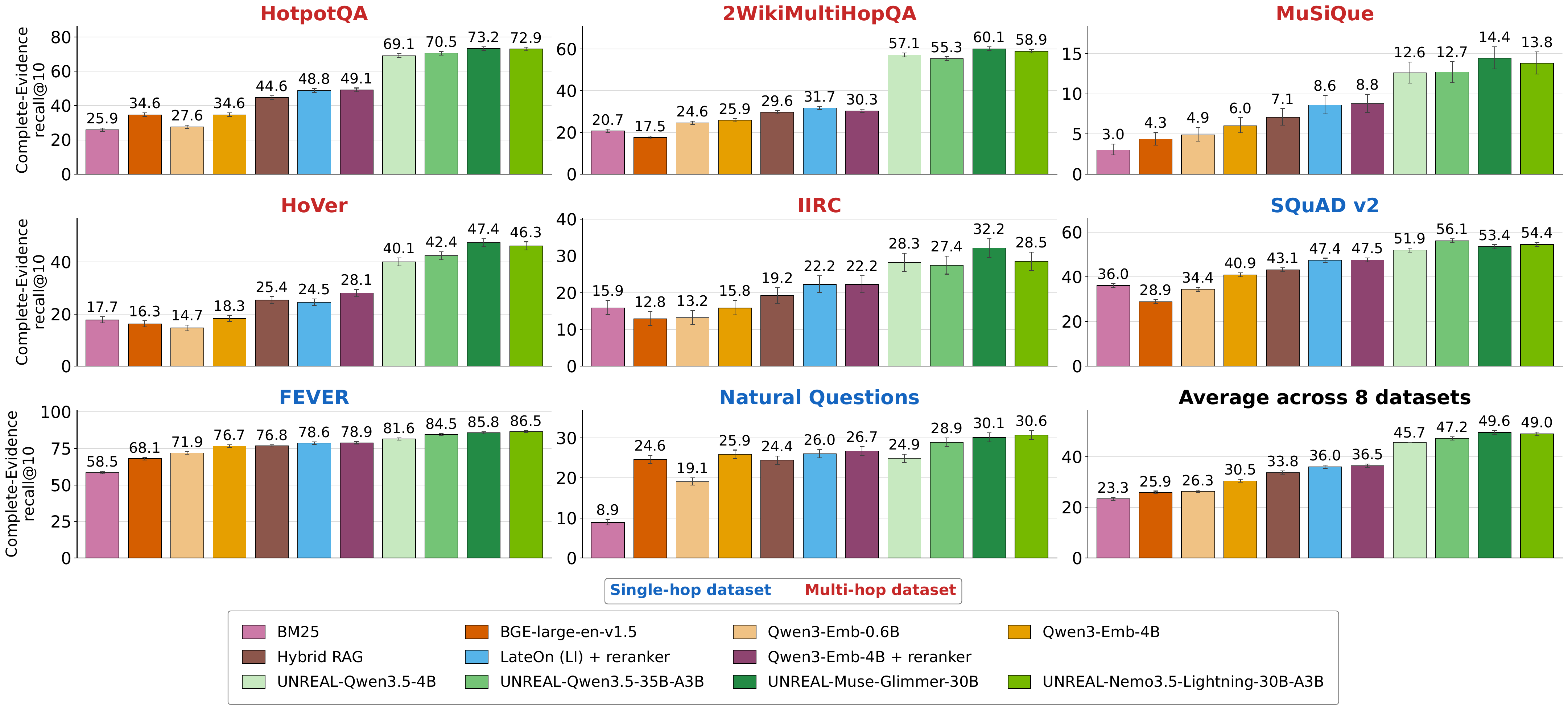}
  \caption{\textbf{UNREAL scales to full-corpus Wikipedia retrieval.} Complete-evidence recall@10 on 21M Wiki-2018 chunks across eight datasets; an example is correct only when all annotated evidence chunks are retrieved. Across four architectures, UNREAL consistently outperforms baselines, especially on multi-hop datasets. Bars show means with 95\% BCa confidence intervals.}
  \vspace{-.2cm}
  \label{fig:intra-all-recall-10}
\end{figure}

\subsection{Results}
We demonstrate UNREAL's architectural generality across four backbones: the dense models Qwen3.5-4B and Muse-Glimmer-30B, the linear-attention hybrid
Qwen3.5-35B-A3B, and the Mamba--attention hybrid
Nemotron-3.5-Lightning-30B-A3B
\citep{meta2026museglimmer,qwen2026qwen35,nvidia2026nemotron35lightning}.

\textbf{Full corpus evaluation.} \cref{fig:intra-all-recall-10} reports complete-evidence recall@10. Each UNREAL backbone leads on most of the eight datasets. The largest
gains occur on HotpotQA, 2WikiMultiHopQA, MuSiQue, and HoVer, challenging multi-hop tasks requiring evidence from multiple articles. A similar pattern
appears for recall@20 in \cref{fig:intra-all-recall-20}. 
For ablation studies please refer to \cref{tab:recipe_ablation}, \cref{tab:corpus_layer_ablation} and \cref{tab:k_ablation}.

\begin{table}[t]
\centering
\small
\caption{Exact match (EM) and macro-averaged F1 on three multi-hop QA datasets, generated by Nemotron-3.5-Lightning under each retrieval method's top-5 retrieved context. UNREAL achieves the highest EM and F1 scores across all three datasets while relying on a single LLM for both retrieval and generation.}
\label{tab:em_f1_datasets}
\begin{tabular}{l cc cc cc}
\toprule
 & \multicolumn{2}{c}{HotpotQA} & \multicolumn{2}{c}{2Wiki} & \multicolumn{2}{c}{MuSiQue} \\
\cmidrule(lr){2-3} \cmidrule(lr){4-5} \cmidrule(lr){6-7}
Method & EM & F1 & EM & F1 & EM & F1 \\
\midrule
No context & 28.5 & 37.7 & 29.9 & 34.6 & 6.8 & 16.6 \\
Oracle & 64.3 & 78.9 & 57.5 & 66.8 & 47.5 & 58.2 \\
\midrule
BM25 & 39.1 & 49.7 & 29.1 & 35.5 & 6.4 & 14.6 \\
BGE-large-en-v1.5 & 39.6 & 51.2 & 28.5 & 35.6 & 8.6 & 17.6 \\
Qwen3-Emb-0.6B & 37.9 & 48.5 & 28.2 & 35.5 & 10.1 & 18.1 \\
Qwen3-Emb-4B & 39.2 & 51.4 & 31.6 & 38.8 & 10.5 & 18.1 \\
Hybrid RAG & 41.5 & 53.1 & 32.5 & 39.8 & 10.0 & 18.9 \\
LateOn + reranker & 43.3 & 55.2 & 30.8 & 38.7 & 11.4 & 20.1 \\
Qwen3-Emb-4B + reranker & 43.3 & 55.3 & 31.4 & 39.1 & 13.2 & 22.6 \\
\midrule
UNREAL-Nemo3.5-Lightning & \textbf{52.6} & \textbf{65.8} & \textbf{42.0} & \textbf{50.1} & \textbf{16.9} & \textbf{25.4} \\
\bottomrule
\end{tabular}
\end{table}

\textbf{Generation evaluation.} \cref{tab:em_f1_datasets} reports end-to-end question-answering performance 
with Nemotron-3.5-Lightning fixed as the generator and only the retriever
providing its top-$5$ chunk context varied. We also report an oracle upper bound using
only oracle chunks and a no-context lower bound. UNREAL attains the highest EM and
F1 on all three datasets while using the same LLM for retrieval and generation. Experiments with  Qwen3.5-35B-A3B as the generator appear in \cref{tab:em_f1_datasets_qwen}.

Overall, UNREAL turns a frozen LLM into a full-corpus retriever by training only the retrieval-token embeddings $\rho$ and layer-mixing weights $\alpha$: fewer than 0.5M added parameters in total, under 0.005\% of each tested backbone's parameter count.


\section{UNREAL as a Long Context Mechanism}
\label{sec:longctx}

In this section, we investigate whether UNREAL's retrieval mechanism can tackle long-context tasks by treating the full prompt as a collection of chunks to retrieve from. 
Many tasks framed as long-context understanding are, in practice, sparse-evidence problems: only a small number of chunks contain the information needed to answer the query, while the remaining context is irrelevant. This structure closely resembles retrieval tasks such as multi-hop question answering \citep{ho2020constructing}.

In this setting, the answer quality depends strongly on retrieving the required evidence. Even when evidence is retrieved, distractors may interfere with generation. Thus, expanding the context window ensures evidence is available, but not that the model can locate and use it effectively. While prior works \citep{xu2024retrievalmeetslongcontext,li2024selfroute,li2025lara} have identified the connection between retrieval and long context data, these approaches rely on an external retriever to select the relevant context. In contrast, UNREAL is the first to use the frozen LLM's internal representations to select evidence and generate the answer.

\vspace{-.1cm}

\subsection{Results}
\label{sec:longctx-results}

Given a long input as context, we treat it as a retrieval problem, i.e., we first split it into $M$ chunks with a mean chunk length of 140 tokens and use the frozen LLM to encode each chunk separately (\cref{eq:method-chunk-rep}).
Then, UNREAL ranks the chunks
against the input query (\cref{eq:method-query-rep,eq:si}), selects the top-$n$ (\cref{eq:intra-background-score}), and
gives their unchanged text to the same frozen LLM for answering (\cref{eq:intrad-generation}). Full experimental details are provided in Appendix~\ref{app:experimental-setting}. Similarly, we include other retrieval baselines, such as BM25 and embedding models paired with rerankers. We wish to emphasize that the UNREAL models used in this section are the same ones presented in \cref{sec:retrieval}, i.e., no special long-context training data was involved.

To evaluate evidence selection at lengths far beyond those of existing
long-context benchmarks, we follow the methodology of LOFT
\citep{lee2024loft} and construct an QA benchmark spanning
8K to 100M tokens. We draw queries and their corresponding oracle chunks
from QA benchmarks and pad them with random distractor passages from
Wiki-2018. As shown in \cref{fig:teaser-figure}, full-context accuracy
drops sharply as context length grows, whereas retrieval-based approaches
perform significantly better. UNREAL outperforms all baselines at large
scales and enables generation using evidence selected from 100M tokens,
far beyond the context limits of current LLMs. For results on standard LOFT see \cref{fig:loft}.

\begin{figure}[t]
  \centering
\includegraphics[width=0.9\linewidth]{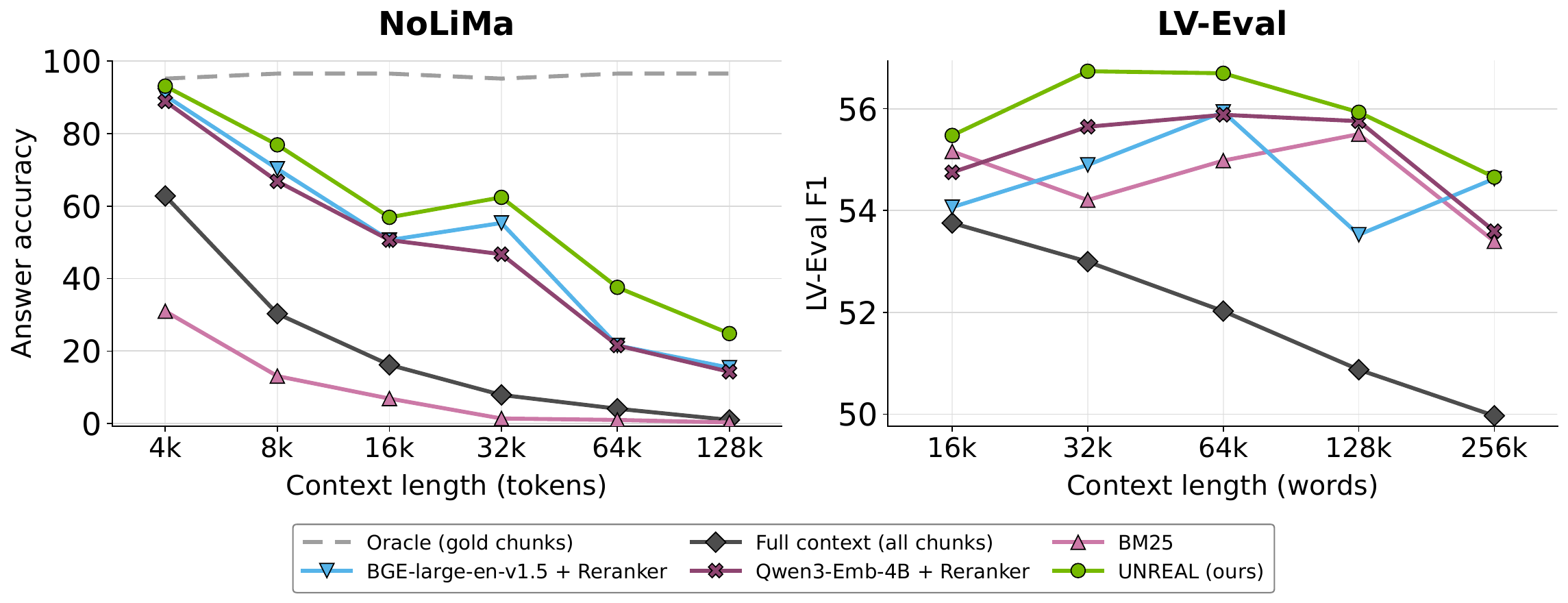}
\caption{\textbf{Left:} NoLiMa accuracy across context lengths (4K--128K tokens) using Nemotron-3-Nano. \textbf{Right:} LV-Eval F1 across context lengths (16K--256K words, averaged over LooGLE-SD and MultiFieldQA-en) using Nemotron-3.5-Lightning. In both cases, full-context performance degrades as context grows, whereas UNREAL achieves the highest performance.}
  \label{fig:longctx-lv-eval-nolima}
\end{figure}

\cref{fig:longctx-lv-eval-nolima} presents results on NoLiMa \citep{modarressi2025nolimalongcontextevaluationliteral}, a needle-in-a-haystack benchmark where queries and target needles share no direct keyword overlap, alongside English QA subsets from LV-Eval \citep{yuan2024lv}. As context lengths scale up to 128K tokens in NoLiMa and 256K words in LV-Eval, full-context performance degrades sharply. In contrast, retrieval baselines remain far more stable, with UNREAL consistently achieving the highest accuracy across all scales.

\begin{figure}[t]
  \centering
\includegraphics[width=0.8\linewidth]{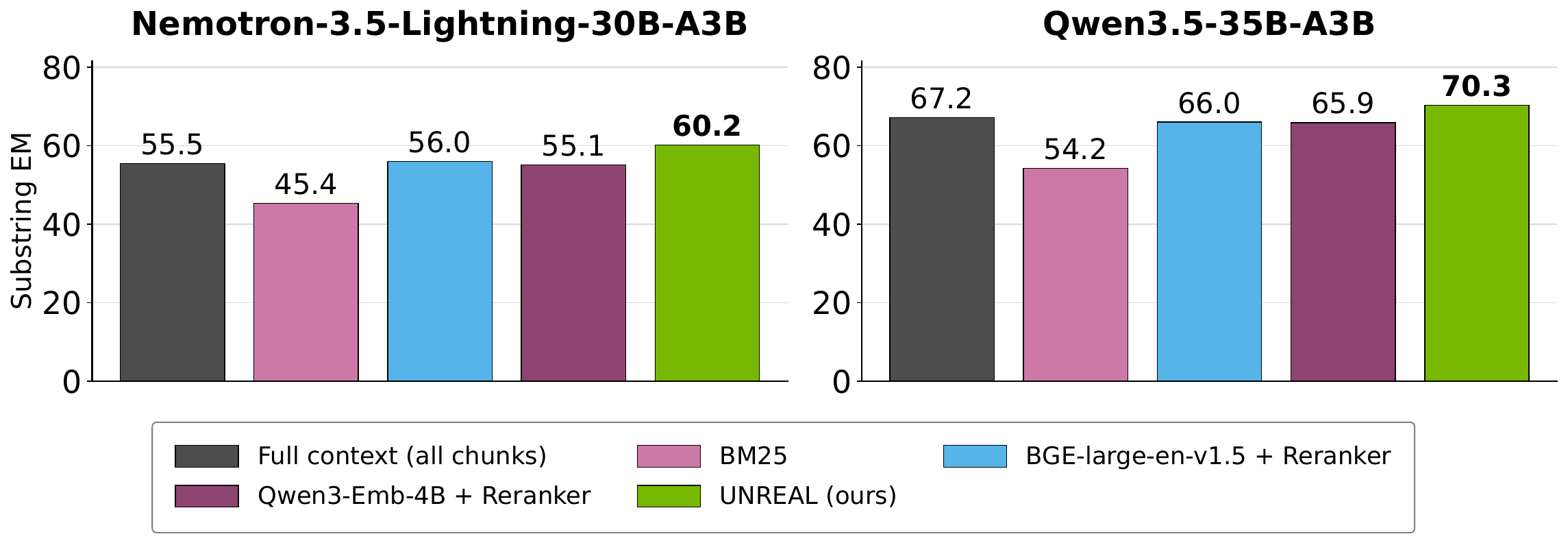}
  \caption{Substring exact-match on the RAG subset of the HELMET long context benchmark, with Nemotron-3.5-Lightning-30B-A3B and Qwen3.5-35B-A3B as generators. UNREAL outperforms full-context inference and other retrieval methods for both generators.}
  \label{fig:longctx-helmet}
\end{figure}

\cref{fig:longctx-helmet} shows results on the RAG subset of the HELMET QA long-context benchmark \citep{yen2024helmetevaluatelongcontextlanguage}. The tasks use context lengths of 8K–128K tokens, and the reported scores are averaged across these lengths. Full-context inference remains a strong baseline, but UNREAL achieves the highest exact-match score for both generators.

\begin{figure}[h]
    \centering
    \includegraphics[width=\linewidth]{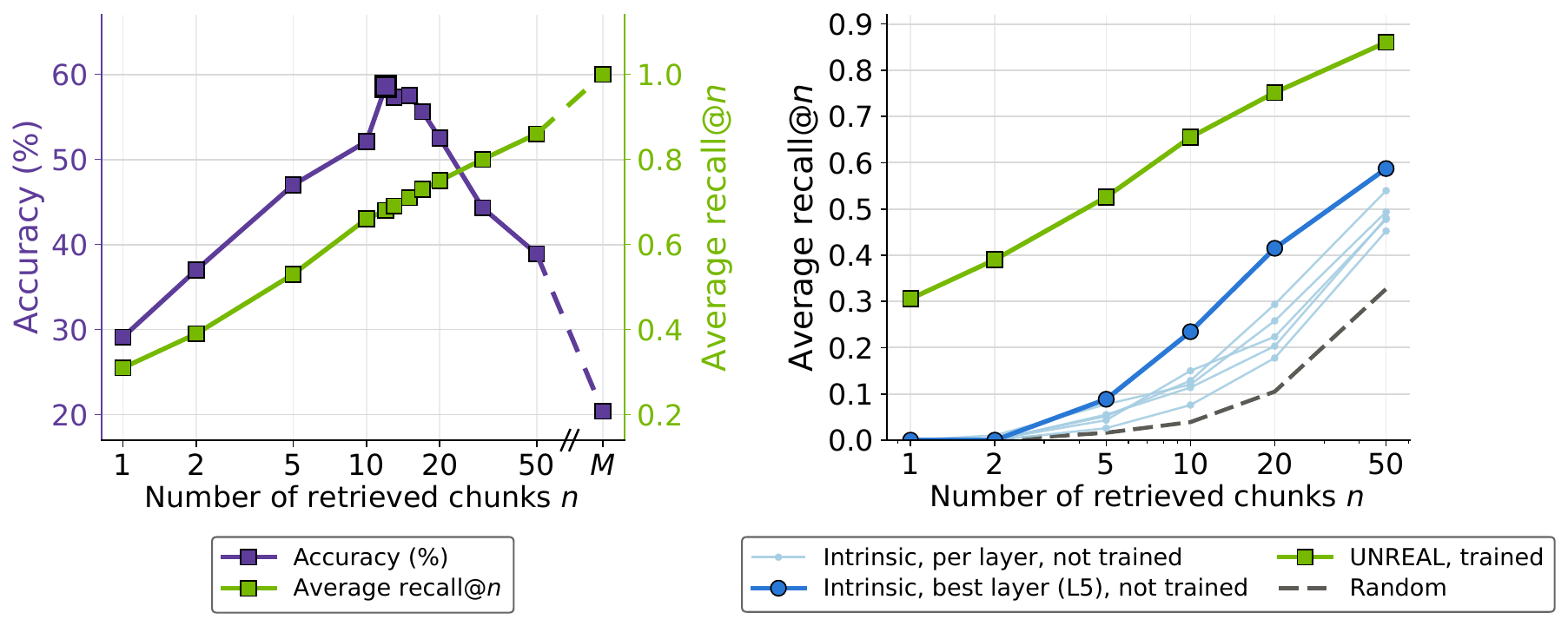}
    \caption{Recall and accuracy on the NoLiMa benchmark for Nemotron-3-Nano.
    \textbf{Left:} The effect of the number of selected chunks $n$ on UNREAL's accuracy and recall.
    \textbf{Right:} Embeddings from the frozen LLM outperform random retrieval, motivating UNREAL, which is trained to enhance this capability.
    }
    \label{fig:intrinsic_recall_and_accuracy_vs_k}
\end{figure}

\textbf{Top-$n$ trade-off.} In \cref{fig:intrinsic_recall_and_accuracy_vs_k} (left) we examine how the number of selected chunks, $n$, affects performance on the long-context NoLiMa benchmark. Accuracy follows an inverted-U pattern: increasing $n$ initially improves performance by raising evidence recall, but eventually reduces performance as additional distractors interfere with generation. Similar non-monotonic trade-offs have been reported in prior work \citep{yu2024rankrag,Jin2024LongContextLM}. UNREAL demonstrates that the same pattern arises when a single decoder-only LLM performs both retrieval and generation.

\textbf{Intrinsic retrieval.} To evaluate whether a frozen LLM can perform retrieval, we test Nemotron-3-Nano on NoLiMa by segmenting the text into chunks. Within every attention layer, we mean-pool the key embeddings to yield a single key vector per chunk. We similarly average the query embeddings over the question tokens and compute a dot product between the pooled key and query vectors. This process yields a per-chunk similarity score, allowing us to compute recall. \cref{fig:intrinsic_recall_and_accuracy_vs_k} (right) shows that these ``intrinsic'' untrained layer representations retrieve evidence above the random baseline. This finding motivates UNREAL, which trains retrieval tokens
$\rho$ to strengthen the retrieval signal already present in the frozen model. For full details, see Appendix~\ref{app:Intrinsic-retrieval}.

\subsection{Efficiency}
\label{sec:longctx-efficiency}

Does UNREAL's additional retrieval pass make it less efficient than
full-context inference, or do its savings dominate at sufficiently long
contexts? Although UNREAL uses separate retrieval and generation passes, both
reduce context-dependent computation. During retrieval, it encodes chunks
independently, avoiding most inter-chunk attention in the prefill. During
generation, it attends only to the selected chunks, greatly reducing both
context attention and the KV-cache size. For fixed chunk size
and selection budget, UNREAL therefore scales linearly with context length,
whereas with standard attention full-context inference scales quadratically.

\begin{figure}[t]
\centering
\includegraphics[
  trim={0 0.8cm 0 0},
  clip,
  width=0.9\linewidth
]{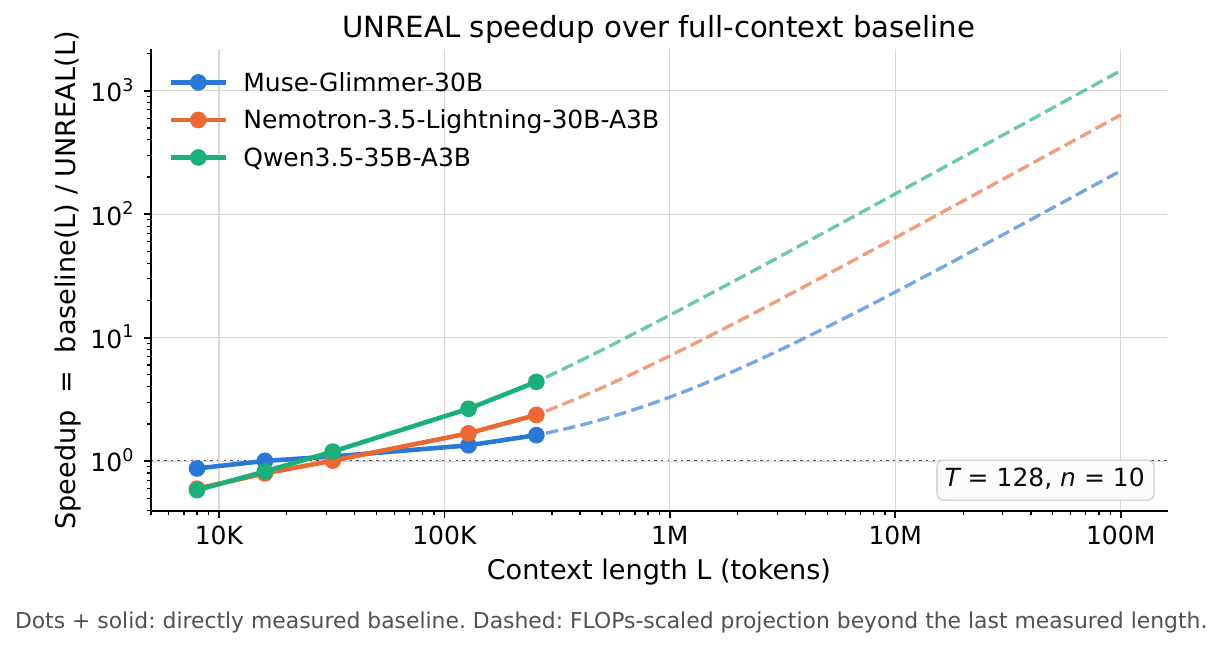}
\caption{Time-to-first-token speedup of UNREAL over full-context inference.
Solid curves use directly measured full-context runtimes; dashed curves use
FLOP-scaled projections beyond the last measured context length.}
\label{fig:unreal-speedup-extrapolated}
\end{figure}

We determine where these savings outweigh the extra pass in both FLOPs and
wall-clock time. Appendix~\ref{app:flops} gives the full FLOP difference in
\eqref{eq:flops-delta}. Applying this to the three tested backbones in our experimental setting, \cref{tab:flops-nstar} shows the UNREAL uses less FLOPs for context lengths over 18K tokens. We also measure stage-wise time-to-first-token with vLLM on a single H100. \Cref{fig:unreal-speedup-extrapolated}
shows the resulting speedup: it grows with context
length across all backbones, with dashed curves projecting the full-context
runtime beyond its last feasible measurement using the FLOP model.
As can be seen, UNREAL reduces time-to-first-token at commonly used context lengths, roughly 32K tokens onward, with substantial speedups at longer contexts.
Full methodology is provided in Appendix~\ref{app:efficiency}.

\section{Related Work}
\label{sec:related}

\textbf{Retrieval models.} ColBERT and ColBERTv2 score passages by
MaxSim over token representations \citep{colbert,colbertv2}. LLM
retrievers contrastively train generators for retrieval
\citep{wang2024e5mistral,ma2024repllama,zhang2025qwen3embedding}. UNREAL
is also contrastively trained, but the chunk encoder and reader remain a frozen generative
decoder. 
REALM, RAG, and related methods
jointly train distinct retriever and reader components
\citep{realm,lewis2021retrievalaugmentedgenerationknowledgeintensivenlp,atlas}.
We instead share the frozen backbone representation across retrieval and generation,
while retaining a separately trained query head.
INTRA \citep{hoffer2026intra} is the closest predecessor: it
introduced retrieval from an encoder--decoder model's internal representations.
We adapt the method to standard decoder-only LLMs and use the selector for
long-context tasks.

\textbf{Efficient attention.} Landmark Attention, InfLLM, Quest,
MoBA, NSA, and HiLS-Attention select blocks of tokens within an LLM's input and perform attention only on those blocks
\citep{mohtashami2023landmark,xiao2024infllm,tang2024quest,lu2025moba,yuan2025nsa,hu2026hilsattention} to improve efficiency.
Sparse attention fixes or learns a restricted set of positions
\citep{beltagy2020longformerlongdocumenttransformer,zaheer2021bigbirdtransformerslonger,kitaev2020reformer,xiao2024streamingllm,xu2026deepseek};
block-level methods summarize and rank KV blocks
\citep{mohtashami2023landmark,xiao2024infllm,tang2024quest,lu2025moba,yuan2025nsa,xu2025xattention};
and hierarchical methods route attention through chunk summaries
\citep{hu2026hilsattention}.
UNREAL instead uses a retrieval-trained mechanism built on the frozen decoder's representations to select chunks before generation, so the LLM receives only the selected text for generation.

\textbf{RAG and long context.} Prior work compares retrieval pipelines with
full-context models \citep{xu2024retrievalmeetslongcontext,li2024selfroute,li2025lara}
or asks whether long context subsumes retrieval \citep{lee2024loft}. 
UNREAL differs from both approaches. Unlike retrieval pipelines, it requires no external retrieval model, avoiding the need to develop and maintain multiple models in production. Unlike full-context baselines, UNREAL is trained on retrieval tasks and removes unselected chunks before generation, preventing them from distracting the generator \citep{shi2023distracted,yu2024defenseofrag}.

\section{Limitations}
Our training data originated from Wikipedia QA, so transfer to
heterogeneous domains and languages remains for future work. For now, our long-context experiments focus on sparse-evidence tasks. UNREAL stores $L_p$
vectors per chunk, costing more to build and store than a single-vector dense index, and relies on a cheap BM25 initial context.
We generate answers with general-purpose LLMs rather than task-specific span extractors such as SpanBERT \citep{joshi2020spanbert}, which are better suited to extractive QA and achieve higher exact match scores. While applying retrieval to long-context tasks substantially reduces context length and improves generation efficiency, it introduces additional steps for chunk encoding and search; these extra steps are inherent to any retrieval-based approach, not unique to UNREAL.

Multi-pass agentic RAG systems interleave reasoning with repeated retrieval \citep{li2025searcho1agenticsearchenhancedlarge,asai2023selfrag}. UNREAL instead focuses on a single-pass retrieval mechanism that could serve as a component within future agentic pipelines. Because these systems evaluate an entire iterative reasoning-and-retrieval pipeline, whereas UNREAL isolates a single retrieval stage, they are not directly comparable and are therefore excluded from our baselines.

\section{Discussion and Future Directions}
\label{sec:conclusion}

This work asks whether corpus retrieval and long-context inference can share a single evidence-selection mechanism inside a decoder-only LLM. UNREAL answers this question by using the LLM’s representations to encode and rank candidate chunks, with fewer than 500K trainable parameters while keeping the backbone frozen. As shown, UNREAL outperformed SOTA retrieval methods on both retrieval and long-context benchmarks. These findings support a unified view of corpus retrieval and long-context inference as the same evidence-selection problem operating at different scales, and provide a path toward LLMs that retrieve and use relevant information without relying on a separate retrieval model.

\subsection{Future directions}

We close by discussing what our main finding, that retrieval supervision improves a model’s evidence selection, implies for training and evaluating context-scalable models.

\textbf{Train the model to recall.} UNREAL makes the model’s internal retrieval ability trainable. Retrieval supervision teaches the model’s representations to rank relevant chunks, improving recall rather than relying on this ability to emerge from pretraining alone. As shown, the learned retriever improves both corpus-scale retrieval and in-prompt long-context selection.

\textbf{An alternative path to context scalability.} These results motivate training in-model retrieval together with generation, rather than treating a larger context window as sufficient. Under this view, sparse-evidence long-context tasks should be formulated primarily as retrieval problems during both training and evaluation. Models should learn to identify the evidence they need before generating, and evaluations should report evidence recall alongside answer accuracy.

\textbf{Benchmarks should separate evidence regimes.} This perspective does
not reduce all long-context understanding to retrieval. New benchmarks should annotate the evidence required for each answer and distinguish sparse-evidence tasks from evidence-dense problems that require integrating information across a large fraction of the input and cannot be solved by selecting a few chunks.

\newpage

\bibliography{main}
\bibliographystyle{iclr2027_conference}
\newpage
\input{appendix}

\end{document}

%% file: math_commands.tex
\usepackage{amsmath,amsfonts,bm}

\def\eqref#1{equation~\ref{#1}}

\def\1{\bm{1}}

\DeclareMathAlphabet{\mathsfit}{\encodingdefault}{\sfdefault}{m}{sl}
\SetMathAlphabet{\mathsfit}{bold}{\encodingdefault}{\sfdefault}{bx}{n}



%% file: appendix.tex
\appendix

\section{Additional experiments}
\label{app:additional-experiments}

\subsection{Corpus-scale retrieval results}
\label{app:corpus-retrieval-results}

\begin{figure}[h]
  \centering
  \includegraphics[width=\linewidth]{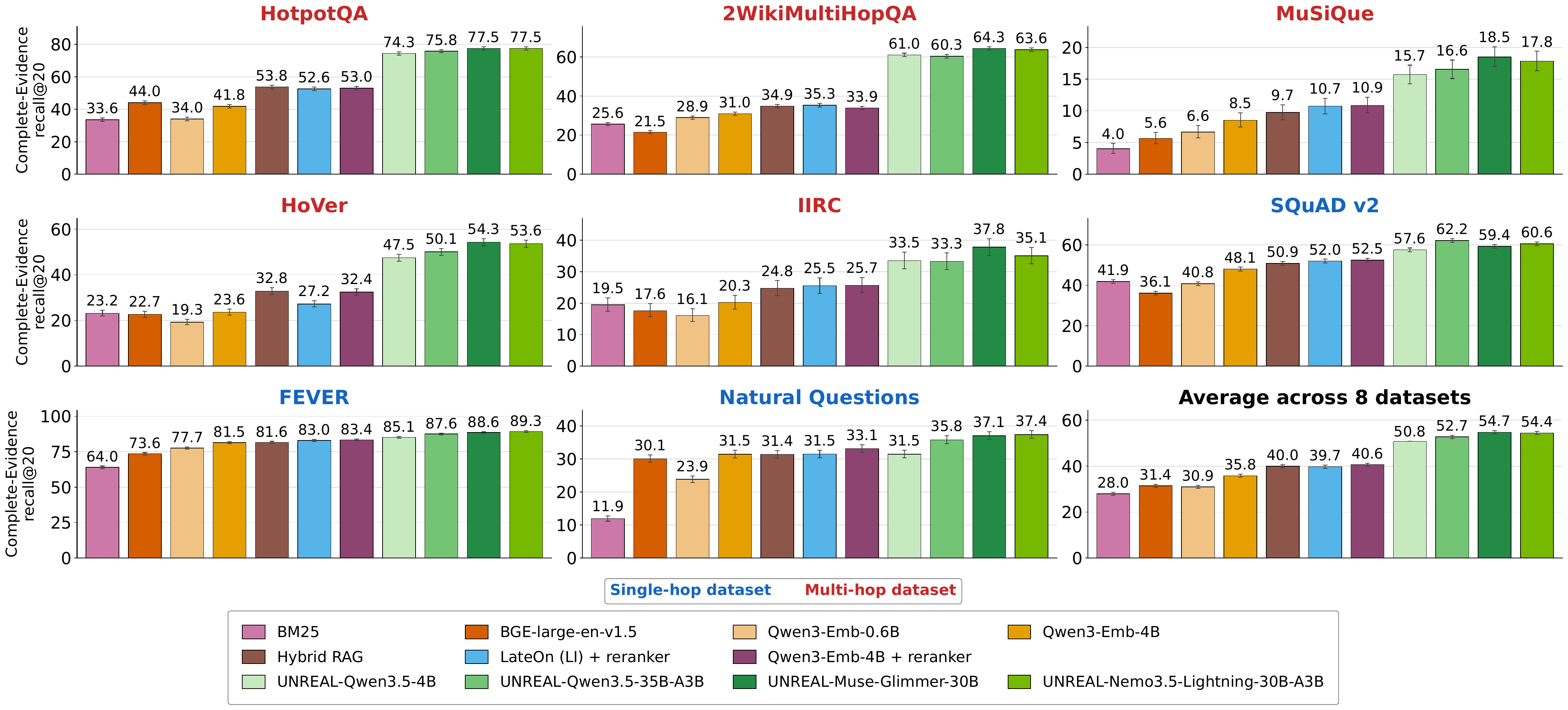}
  \caption{\textbf{Complete-evidence recall@20} on 21M Wiki-2018 chunks across 8 datasets; an example
  is correct only when all annotated evidence chunks are retrieved. Bars show means with 95\% BCa confidence intervals.}
  \label{fig:intra-all-recall-20}
\end{figure}

\begin{figure}[h]
  \centering
  \includegraphics[width=\linewidth]{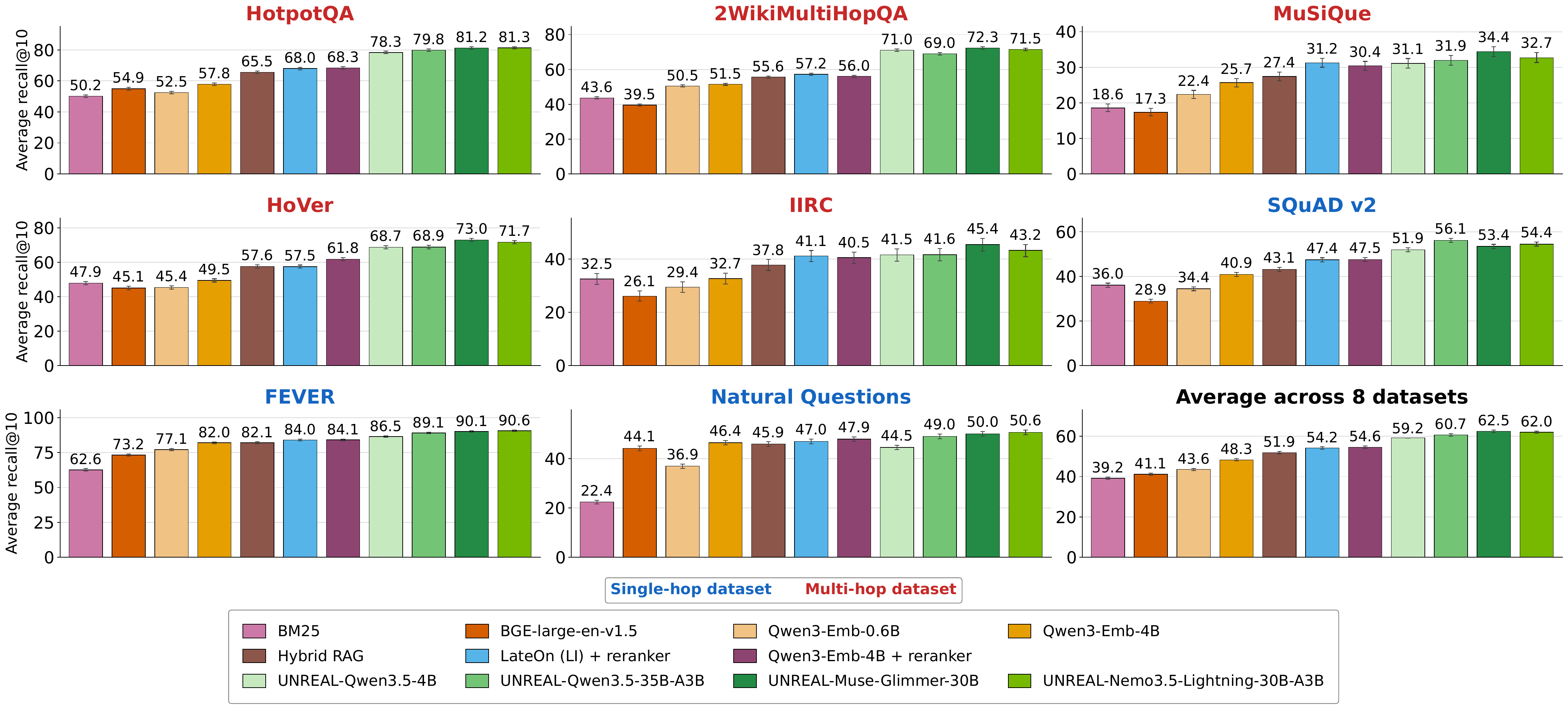}
  \caption{\textbf{Average recall@10} on 21M Wiki-2018 chunks across 8 datasets. Bars show
  means with 95\% BCa confidence intervals.}
  \label{fig:intra-avg-recall-10}
\end{figure}

\begin{figure}[h]
  \centering
  \includegraphics[width=\linewidth]{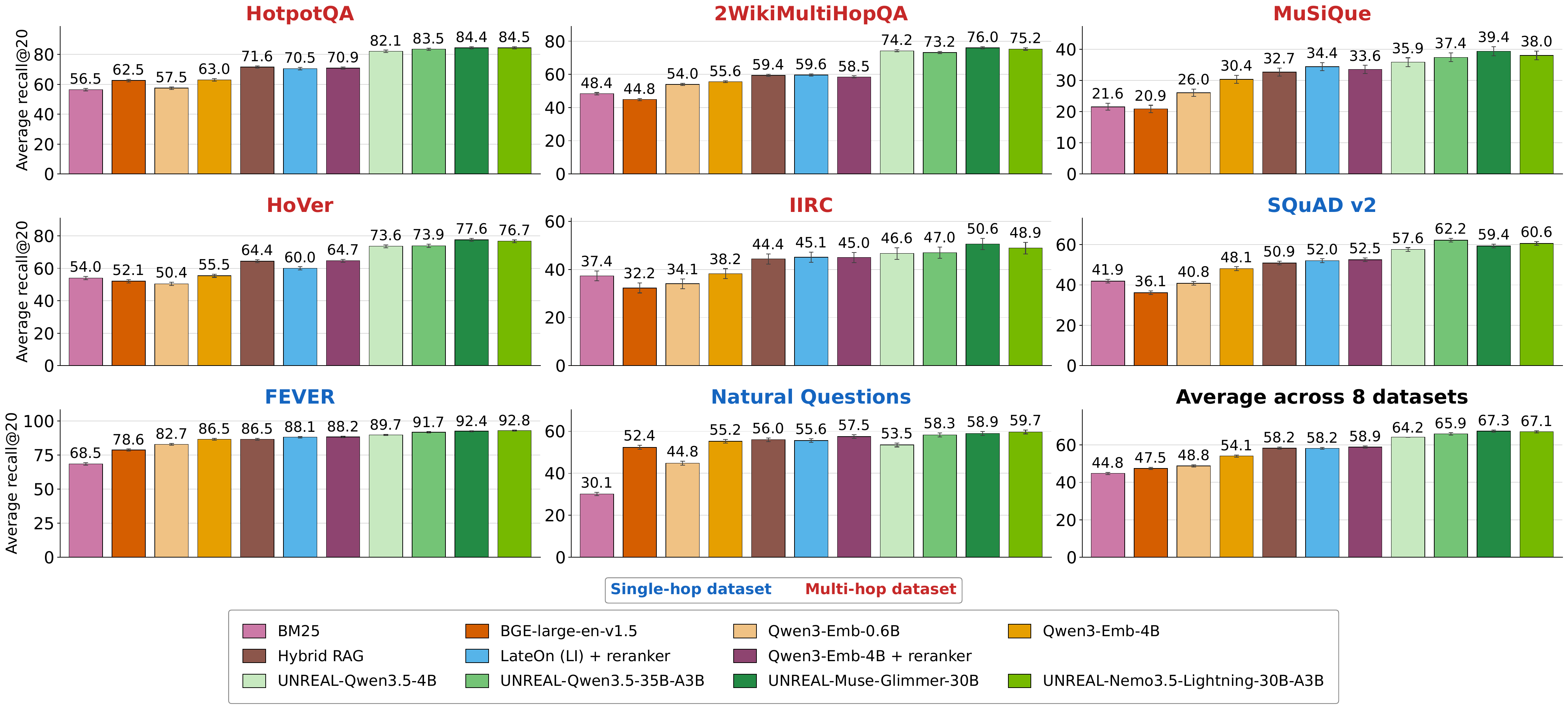}
  \caption{\textbf{Average recall@20} on 21M Wiki-2018 chunks across 8 datasets. Bars show
  means with 95\% BCa confidence intervals.}
  \label{fig:intra-avg-recall-20}
\end{figure}

\cref{fig:intra-avg-recall-10,fig:intra-avg-recall-20,fig:intra-all-recall-20} report Complete-evidence recall@20, average recall@10, and average recall@20, respectively, across 8 datasets using the full Wiki-2018 corpus, demonstrating UNREAL's superiority over SOTA embedding models, rerankers, and late-interaction methods.

\vspace{4cm}
\cref{tab:em_f1_datasets_qwen} reports end-to-end question-answering performance 
with Qwen3.5-35B-A3B fixed as the generator and only the retriever
providing its top-$5$ chunk context varied.

\vspace{2cm}
\begin{table}[h]
\centering
\caption{Exact match (EM) and macro-averaged F1 on three QA datasets, generated by Qwen3.5-35B-A3B under each retrieval method's top-5 retrieved context.
``Oracle'' denotes the upper-bound result, obtained by answering the question based only on the oracle chunks.
}
\label{tab:em_f1_datasets_qwen}
\begin{tabular}{l cc cc cc}
\toprule
 & \multicolumn{2}{c}{HotpotQA} & \multicolumn{2}{c}{2Wiki} & \multicolumn{2}{c}{MuSiQue} \\
\cmidrule(lr){2-3} \cmidrule(lr){4-5} \cmidrule(lr){6-7}
Method & EM & F1 & EM & F1 & EM & F1 \\
\midrule
No context & 25.0 & 34.6 & 27.7 & 33.3 & 6.9 & 14.8 \\
Oracle & 60.8 & 74.9 & 53.6 & 61.7 & 42.6 & 54.6 \\
\midrule
BM25 & 36.8 & 47.3 & 28.1 & 34.4 & 7.0 & 15.2 \\
BGE-large-en-v1.5 & 36.9 & 47.3 & 27.6 & 33.7 & 9.7 & 18.1 \\
Qwen3-Emb-0.6B & 34.9 & 46.0 & 28.5 & 35.0 & 7.7 & 15.4 \\
Qwen3-Emb-4B & 36.5 & 48.7 & 31.7 & 38.4 & 10.4 & 18.4 \\
Hybrid RAG & 39.3 & 51.3 & 30.7 & 37.5 & 8.9 & 18.1 \\
LateOn + reranker & 41.3 & 53.2 & 31.5 & 38.5 & 10.8 & 19.6 \\
Qwen3-Emb-4B + reranker & 40.8 & 52.6 & 31.2 & 37.9 & 11.8 & 20.8 \\
\midrule
UNREAL-Qwen3.5-35B-A3B & \textbf{48.2} & \textbf{61.6} & \textbf{40.2} & \textbf{47.2} & \textbf{17.8} & \textbf{27.1} \\
\bottomrule
\end{tabular}
\end{table}

\subsection{Long-context results}
\label{app:long-context-results}

\begin{figure}[h]
  \centering
  \includegraphics[width=0.6\linewidth]{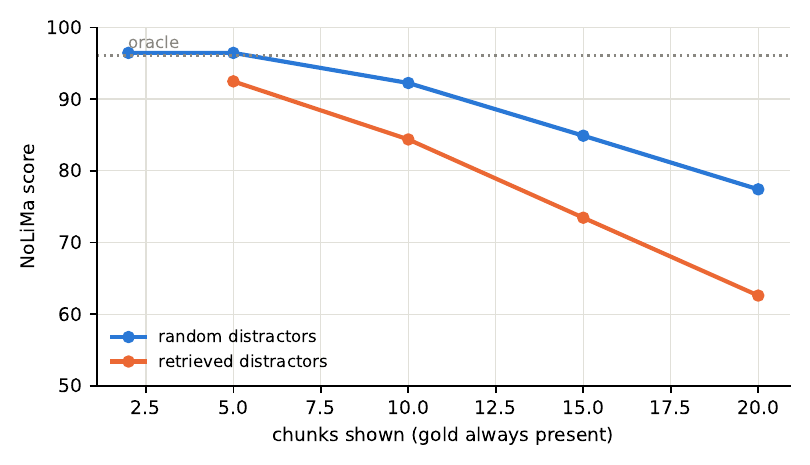}
\caption{\textbf{Distractor identity matters.} With the gold chunk always present, retrieved distractors cause more interference than the same number of random distractors, and the gap grows with the budget.}
\label{fig:dilution}
\end{figure}

In \cref{fig:dilution} we present the effect of retrieved distractors vs random distractors on the NoLiMa dataset. As can be seen, retrieved distractors reduce the NoLiMa score more than random distractors.

\begin{figure}[t]
  \centering
  \includegraphics[width=\linewidth]{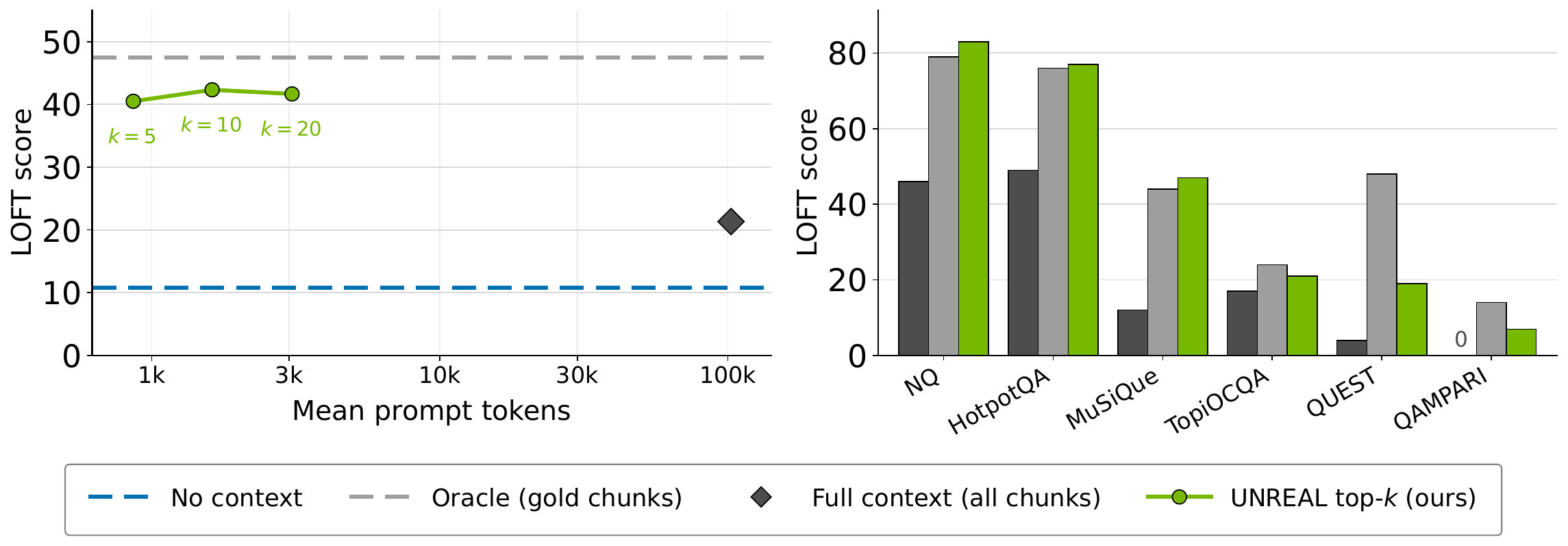}
  \caption{\textbf{LOFT generation.} (a) All selective budgets outperform
  full-context reading. (b) Performance by dataset at $n=10$ against gold-only
  context. We use each dataset's designated LOFT metric.}
  \label{fig:loft}
\end{figure}

\newpage

\subsubsection{Intrinsic retrieval}
\label{app:Intrinsic-retrieval}

\cref{fig:intrinsic_recall_and_accuracy_vs_k} (right) evaluates whether the frozen Nemotron-3-Nano can retrieve evidence on NoLiMa without retrieval training. We split each context into chunks and score them against the question using the model's attention keys and queries.

For attention layer $\ell$ and head $h$, let $k_{i,t}^{\ell,h}$ denote the key vector at token $t$ of chunk $c_i$, and $q_t^{\ell,h}$ the query vector at token $t$ of the question $x$. We mean-pool these vectors over the chunk and question tokens, respectively:
\[
\bar{k}_i^{\ell,h} = \frac{1}{|c_i|}\sum_{t=1}^{|c_i|} k_{i,t}^{\ell,h},
\qquad
\bar{q}^{\ell,h} = \frac{1}{|x|}\sum_{t=1}^{|x|} q_t^{\ell,h}.
\]
We compute their dot product within each head and average over the $H_\ell$ heads to obtain a chunk score:
\[
s_i^\ell(x) = \frac{1}{H_\ell}\sum_{h=1}^{H_\ell}
\left\langle \bar{q}^{\ell,h}, \bar{k}_i^{\ell,h} \right\rangle.
\]
For each layer separately, we rank chunks by $s_i^\ell(x)$ and compute recall@$n$ over the top-$n$ chunks. No retrieval tokens or learned layer weights are used. These untrained representations retrieve evidence above the random baseline.

\subsection{Ablation studies}
\label{app:ablation_studies}

\cref{tab:recipe_ablation} presents a recipe ablation for Nemotron-3.5-Lightning-30B-A3B, in which we vary, one at a time, the design choices used in the main method. \cref{tab:corpus_layer_ablation} presents an ablation study of the layer from which corpus chunk embeddings are produced. We find that middle-to-late layers yield the strongest retrieval performance, a result that is reproduced across other architectures as well.

\begin{table*}[h]
    \centering
    \small
    \caption{Recipe ablation for UNREAL-Nemotron-3.5-Lightning-30B-A3B, single-knob departures from the main method. $R$ is the number of retrieval tokens; $G$ is the number of groups they are mean-pooled into on the query side; ML (multi-layer) vs.\ SL (single-layer) is whether the read-out combines all 6 global-attention blocks or only the block used for the corpus readout; HN is the number of hard negatives per query; $|C_0(x)|$ is the number of BM25-retrieved context chunks prepended to the query; $P$ is the number of learnable soft prompt tokens used at the beginning of the sequence.
    Avg.\ R@10 is recall averaged over each query's oracle chunks; All R@10 requires every oracle chunk to be retrieved.}
    \label{tab:recipe_ablation}
    \begin{tabular}{lccccc}
        \toprule
        & \multicolumn{2}{c}{HotpotQA} & \multicolumn{2}{c}{MuSiQue} & \multicolumn{1}{c}{SQuAD v2} \\
        \cmidrule(lr){2-3} \cmidrule(lr){4-5} \cmidrule(lr){6-6}
        Design choice & Avg R@10 & All R@10 & Avg R@10 & All R@10 & Avg R@10 \\
        \midrule
        Baseline & 81.28 & 72.93 & 32.70 & 13.79 & 54.43 \\
        \midrule
        $R\!=\!64 \to 16$ & 77.93 \drop{-3.35} & 67.83 \drop{-5.10} & 31.91 \drop{-0.79} & 13.04 \drop{-0.75} & 53.82 \drop{-0.61} \\
        $G\!=\!4 \to 1$ & 77.91 \drop{-3.37} & 68.37 \drop{-4.56} & 29.69 \drop{-3.01} & 11.49 \drop{-2.30} & 52.94 \drop{-1.49} \\
        ML $\to$ SL & 73.02 \drop{-8.26} & 61.01 \drop{-11.92} & 26.20 \drop{-6.50} & 9.86 \drop{-3.93} & 46.93 \drop{-7.50} \\
        HN $\!=\!500 \to 100$ & 79.40 \drop{-1.88} & 70.59 \drop{-2.34} & 28.77 \drop{-3.93} & 11.58 \drop{-2.21} & 52.86 \drop{-1.57} \\
        $|C_0(x)| \!=\!5 \to 1$ & 78.50 \drop{-2.78} & 68.83 \drop{-4.10} & 31.36 \drop{-1.34} & 11.95 \drop{-1.84} & 52.68 \drop{-1.75} \\
        $|C_0(x)| \!=\!5 \to 0 $ & 71.08 \drop{-10.20} & 56.73 \drop{-16.20} & 28.83 \drop{-3.87} & 8.94 \drop{-4.85} & 50.16 \drop{-4.27} \\
        $P\!=\!1 \to 0$ & 79.61 \drop{-1.67} & 70.86 \drop{-2.07} & 32.17 \drop{-0.53} & 13.21 \drop{-0.58} & 52.92 \drop{-1.51} \\
        $L_p \!=\!7 \to 1$ & 78.20 \drop{-3.08} & 69.11 \drop{-3.82} & 30.28 \drop{-2.42} & 12.45 \drop{-1.34} & 48.52 \drop{-5.91} \\
        \bottomrule
    \end{tabular}
\end{table*}

\begin{table*}[h]
    \centering
    \small
    \caption{Ablation on the layer used to create the corpus chunk
    embeddings, for UNREAL-Nemotron-3.5-Lightning-30B-A3B.}
    \label{tab:corpus_layer_ablation}
    \begin{tabular}{lccccc}
        \toprule
        & \multicolumn{2}{c}{HotpotQA} & \multicolumn{2}{c}{MuSiQue} & \multicolumn{1}{c}{SQuAD v2} \\
        \cmidrule(lr){2-3} \cmidrule(lr){4-5} \cmidrule(lr){6-6}
        Corpus layer & Avg R@10 & All R@10 & Avg R@10 & All R@10 & Avg R@10 \\
        \midrule
        Layer 8  & 79.68 & 70.25 & 31.74 & 12.33 & 53.66 \\
        Layer 18 & 78.98 & 69.42 & 31.65 & 12.45 & 54.05 \\
        Layer 30 & 80.19 & 71.39 & 32.40 & 13.33 & 54.85 \\
        Layer 38 & 80.62 & 71.99 & 31.94 & 12.29 & 54.01 \\
        Layer 42 & \textbf{81.28} & \textbf{72.93} & 32.70 & \textbf{13.79} & \textbf{54.43} \\
        Layer 48 & 80.19 & 71.64 & \textbf{32.78} & 12.75 & 54.13 \\
        \bottomrule
    \end{tabular}
\end{table*}

\cref{tab:k_ablation} examines how the number of pooled vectors per corpus chunk, $L_p$, affects retrieval quality for Nemotron-3.5-Lightning-30B-A3B. We encode the Wiki-2018 corpus at $L_p \in {1, 3, 7, 11}$ and report Average recall@10 and Complete-evidence recall@10. Recall increases monotonically with $L_p$ across both metrics, confirming that a finer-grained chunk representation yields more accurate retrieval, though this comes at the cost of higher memory and compute during evaluation.

\begin{table*}[t]
    \centering
    \small
    \caption{Effect of the number of pooled vectors per corpus chunk $L_p$ on retrieval recall@10, for UNREAL-Nemotron-3.5-Lightning-30B-A3B. Avg.\ R@10 is the recall averaged over each query's oracle chunks; All R@10 requires every oracle chunk of a query to be retrieved. Recall increases monotonically with $L_p$, but at the cost of higher memory and compute during evaluation.}
    \label{tab:k_ablation}
    \begin{tabular}{lccccccc}
        \toprule
        & \multicolumn{2}{c}{HotpotQA} & \multicolumn{2}{c}{MuSiQue} & \multicolumn{2}{c}{2Wiki} & SQuAD v2 \\
        \cmidrule(lr){2-3} \cmidrule(lr){4-5} \cmidrule(lr){6-7} \cmidrule(lr){8-8}
        $L_p$ & Avg R@10 & All R@10 & Avg R@10 & All R@10 & Avg R@10 & All R@10 & Avg R@10 \\
        \midrule
        1  & 78.20 & 69.11 & 30.28 & 12.45 & 67.43 & 55.21 & 48.52 \\
        3  & 79.34 & 70.58 & 31.15 & 12.70 & 69.51 & 57.11 & 51.61 \\
        7  & 81.28 & 72.93 & 32.70 & 13.79 & 71.52 & 58.90 & 54.43 \\
        11 & 81.54 & 73.40 & 34.34 & 14.67 & 72.38 & 59.68 & 55.42 \\
        \bottomrule
    \end{tabular}
\end{table*}

\section{Experimental setting}
\label{app:experimental-setting}

\subsection{Retrieval data}
\label{app:retrieval-data-setting}

We construct our retrieval data setting by mapping gold evidence from existing datasets to a shared DPR Wikipedia 2018 corpus \citep{karpukhin2020dense}, using a KILT-like method \citep{petroni2021kilt}. The corpus consists of around 21M chunks, with a median of around 141 Nemotron-3.5-Lightning tokens per chunk. The downstream tasks span three categories: single-hop QA (Natural Questions \citep{kwiatkowski2019natural}, SQuAD v2 \citep{rajpurkar2018squadv2}), multi-hop QA (HotpotQA \citep{yang2018hotpotqa}, 2WikiMultiHopQA \citep{ho2020constructing}, MuSiQue \citep{trivedi2022musique}, IIRC \citep{ferguson2020iirc}), and fact verification (FEVER \citep{thorne2018fever}, HoVer \citep{jiang2020hover}).

Based on KILT \citep{petroni2021kilt}, for each dataset, we map its oracle chunks to Wikipedia 2018 chunks via title matching. We then compute a text-alignment quality score based on length-robust unigram, bigram, and trigram containment overlap between the oracle evidence text and each candidate chunk. The score ranges from 0 to 1, where 1.0 means that the oracle text is essentially contained verbatim in the chunk. For NQ, we directly reuse existing DPR bi-encoder gold mappings, bypassing this KILT mapping step. For evaluation, similar to KILT, we report results on a filtered set that retains an oracle only if it comes from an exact title match or has a fallback alignment score above a quality threshold (0.6 by default).

We additionally retrieve lexical signals for each query. For each query, we use its input text as a BM25 query and retain the top chunks as BM25 context, which is provided as additional input to UNREAL ($C_0(x)$). We then mine oracle-seeded hard negatives: the query's oracle chunks were used as seed queries by running their text through BM25. The top lexically similar chunks across all seeds are pooled and deduplicated, and any chunk overlapping with the query's oracle set is discarded. This yields up to 512 hard negatives per query that are lexically close to the gold evidence but guaranteed not to belong to it, providing a clean hard-negative pool for contrastive training.

\subsection{Long-context benchmarks}
\label{app:long-context-benchmarks}

We wish to note that the UNREAL models evaluated in the long-context setting were not trained on long-context data. Instead, the UNREAL modules were trained exclusively on standard retrieval data (as detailed in \cref{app:retrieval-data-setting}) and subsequently evaluated on both retrieval and long-context tasks.

Following the methodology of LOFT~\citep{lee2024loft}, we construct a new open-domain QA benchmark scaling context lengths from 8K to 100M tokens (\cref{fig:teaser-figure}). Our goal is to evaluate long-context generation on context scales significantly larger than those present in current benchmarks.

We sample 500 queries at random from the validation splits of four QA datasets: Natural Questions~\citep{kwiatkowski2019natural}, HotpotQA~\citep{yang2018hotpotqa}, 2WikiMultiHopQA~\citep{ho2020constructing}, and MuSiQue~\citep{trivedi2022musique}, totaling 2,000 questions. Each query is paired with oracle passages from the DPR Wiki-2018 corpus (approximately 21 million passages). We retain only questions whose gold passages match corpus entries by exact article title.

To construct the extended contexts, all queries share a single distractor pool. We remove all gold passages associated with any validation query across the four datasets to ensure distractors do not overlap with target evidence. Contexts are constructed across ten lengths: 8K, 16K, 32K, 128K, 256K, 512K, 1M, 10M, 30M, and 100M tokens, measured using the Nemotron tokenizer over context passages. Distractors are drawn sequentially from the head of the shuffled corpus to fill the token budget alongside the gold passages. Each gold passage is placed at a relative context depth sampled uniformly at random for that query; these relative depths remain fixed across all context lengths. Consequently, as context length scales, a question maintains its gold evidence at identical relative positions, altering only the volume of surrounding distractor text. All stochastic choices, including question sampling, corpus shuffling, and passage positioning, use fixed random seeds to guarantee deterministic inputs across evaluated models and runs.

To evaluate retrieval methods on long-context benchmarks, contexts are split into sentence-aligned chunks averaging 141 tokens (mostly ranging from 123 to 165 tokens) to match the Wiki-2018 length distribution used during retriever training. The chunks form an exact cover of the source text and are created on a per-document basis to prevent them from crossing document boundaries. For generation, retrieved chunks are reassembled into native prompt templates in their original document order and scored using native metrics.

\subsection{Method and training configuration}
\label{app:training-configuration}

On the corpus side, we use the full Wikipedia-2018 corpus, containing around 21M chunks. Each chunk is encoded only once using the frozen LLM, by extracting the outputs of a single layer and compressing them into $L_p=7$ pooled vectors.

On the query side, we use $R=64$ retrieval tokens pooled into $G=4$ groups, together with an initial context $C_0(x)$ consisting of the top-5 BM25 chunks. The query embeddings are obtained by a learnable sum of the full-attention outputs of the model. Moreover, only for query processing, a single learned \emph{soft-prompt} vector is concatenated to the beginning of the retrieval input. Unlike $\rho$, this vector is never read out or scored; it only steers the frozen LLM toward a retrieval-oriented computation and is trained jointly with $\rho$ and $\alpha$.
These components are the only learnable parameters in UNREAL, 
totaling fewer than 500K parameters, or less than $2\times 10^{-5}$ of the frozen model in all cases.
The exact number of learnable parameters is $(R+1)D + |\alpha|$, where $D$ is the embedding dimension and $|\alpha|$ is the number of layer-mixing weights.

We trained the UNREAL models using the training sets of Natural Questions \citep{kwiatkowski2019natural}, SQuAD v2 \citep{rajpurkar2018squadv2}, HotpotQA \citep{yang2018hotpotqa}, 2WikiMultiHopQA \citep{ho2020constructing}, MuSiQue \citep{trivedi2022musique}, IIRC \citep{ferguson2020iirc}, FEVER \citep{thorne2018fever}, and HoVer \citep{jiang2020hover}, all mapped into a single shared Wikipedia-2018 corpus (see Appendix~\ref{app:retrieval-data-setting}).
We note that all retrieval baselines used in this work, such as Qwen3-Embedding \citep{zhang2025qwen3embedding}, were also fine-tuned on high-quality retrieval datasets like HotpotQA and Natural Questions.

For the contrastive objective, we draw $500$ hard negatives per query from a BM25 ranking seeded on that query's oracle chunks. We discard the top $10$ ranks, as they are typically near-duplicates of the oracle chunks themselves. $\mathcal{B}(x)$ is then the deduplicated union of the oracle and hard-negative chunks from all queries in a microbatch, ensuring that a chunk serving as an oracle for one query is never treated as a negative for that query. We train for $15K$ steps with a global batch of $256$ queries, using AdamW with $\beta=(0.9,0.95)$, weight decay $0.1$, gradient clipping at norm $1.0$, and $100$ warmup steps followed by linear decay to zero. At evaluation time, each query is scored exhaustively against all $M$ chunks, without an approximate index or candidate pre-filter.

The per-backbone settings are as follows. For Qwen3.5-35B-A3B ($40$ blocks, $d=2048$), we read the index at $\ell_c=27$, using a learning rate of $5\times 10^{-3}$. For Muse-Glimmer-30B ($52$ blocks, $d=6656$), we use $\ell_c=47$, and a learning rate of $5\times 10^{-3}$. For Nemotron-3.5-Lightning-30B-A3B ($52$ blocks, $d=2688$), we use $\ell_c=42$, and a learning rate of $10^{-2}$.

\section{FLOP Analysis}
\label{app:flops}

UNREAL replaces full-context inference with chunk encoding, retrieval, and
generation over selected evidence. This avoids processing irrelevant context
but adds retrieval and re-encoding work. We compare these costs and derive the
context length above which UNREAL requires fewer FLOPs, counting only the work
on which the two pipelines differ.

\paragraph{Full-context inference vs.~UNREAL.}
Full-context inference prefills $[\text{context}, x]$. UNREAL instead:
\begin{enumerate}
\item[(i)] Encodes each chunk independently up to layer $\ell_c$
(\eqref{eq:method-chunk-rep});
\item[(ii)] Runs one forward pass through all $L$ layers on $x_{\text{ret}}$
(\eqref{eq:x-rho}), of length
$|x_{\text{ret}}| = |C_0(x)|\,T + T_q + R$;
\item[(iii)] Prefills $[C_{\text{UNREAL}}(x), x]$, of length $nT + T_q$, for
generation (\eqref{eq:intrad-generation}).
\end{enumerate}

After prefill, both pipelines perform the same autoregressive decoding except
for the cached context length: full context attends to $N+T_q$ prompt tokens,
whereas UNREAL attends to $nT+T_q$. Thus UNREAL uses fewer attention FLOPs and a smaller KV cache, which benefits memory-bound decoding.

\subsection{Setting}
All chunks have equal length, $T_i \equiv T$, so the context given to the model contains
$N \equiv MT$ tokens. Both pipelines use a KV cache, and $T_g$ denotes the
number of decode forward passes.
A multiply-add counts as two FLOPs.

\paragraph{Weights.} For identical layers, passing one token through the
first $\ell$ layers costs $2\Theta\ell/L$ FLOPs, where $\Theta$ counts
backbone matrix parameters active per token. It includes all sequence-mixer
and dense-FFN projections; in MoE layers, it counts the router, shared
experts, and selected routed experts. It excludes token embeddings, which are
lookups, and the LM output head, whose generation cost cancels in \eqref{eq:flops-delta}.

\paragraph{Attention.} One causal query-key pair costs $4L_a d_a$ FLOPs, summed
over the $L_a$ global-attention layers: $2d_a$ for the score and $2d_a$ for
the value accumulation. For a standard decoder, $L_a = L$ and $d_a = d$.

Linear-attention and state-space layers have no position-dependent pairwise
term, so we include their projections in $\Theta$ and neglect their small
recurrent-update cost. We also omit sliding-window attention: full context
attends to up to $W$ keys per token, so this conservatively understates
UNREAL's savings.

\paragraph{Omitted operations.} Softmax, normalization, and activations are
ignored. The selection stage is also neglected: BM25 for $C_0(x)$ involves
no dense matrix products, and the remaining selection steps are tiny. These
are mean pooling (about $d$ FLOPs per token), the $\alpha$-weighted layer
sum, and MaxSim scoring (\eqref{eq:si}, $2dGL_p$ FLOPs per chunk).
Chunk encoding alone costs $2\Theta\ell_c/L$ per context token, and the ratio
$(1 + 2GL_p/T)\,dL/(2\Theta\ell_c)$ is about $2\times10^{-7}$ for the models
below (this ratio covers pooling and MaxSim). 

\subsection{FLOP difference: UNRAEL vs.~full-context inference}
The two pipelines perform identical work on two things:
\begin{itemize}
\item the context tokens in layers $1,\dots,\ell_c$, including all
intra-chunk attention there;
\item the shared generation-side work: the query tokens' weight products and
query-to-query attention in the prefill; all decode-step weight products and
attention to the query and previously generated tokens; and all required
generation logits.
\end{itemize}

Let $\lambda$ denote the fraction of $\Theta$ in layers
$\ell_c+1,\dots,L$, for identical layers, $\lambda = 1 - \ell_c/L$. Let
$L_{a,\leq c}$ count global-attention layers through $\ell_c$. The difference
$\Delta \equiv F_{\text{full}} - F_{\text{UNREAL}}$ is
\begin{equation}
\begin{split}
\Delta ={}&
\underbrace{2\Theta\lambda N}_{\text{(a)}}
+ \underbrace{2L_a d_a\bigl[N(N-T) + 2(T_q+T_g)(N-nT)\bigr]}_{\text{(b)}} \\
&- \underbrace{\bigl[2\Theta\,nT + 2L_a d_a nT(nT+1)\bigr]}_{\text{(c)}} \\
&- \underbrace{\bigl[2\Theta\,|x_{\text{ret}}| + 2L_a d_a
|x_{\text{ret}}|(|x_{\text{ret}}|+1)\bigr]}_{\text{(d)}} \\
&+ \underbrace{2d_a(L_a-L_{a,\leq c})N(T+1)}_{\text{(e)}}.
\end{split}
\label{eq:flops-delta}
\end{equation}

\begin{itemize}
\item[(a)] \emph{Deeper layers on the context (UNREAL saving).} Full-context
inference passes every context token through all $L$ layers. UNREAL reads
the chunk embeddings at layer $\ell_c$ (\eqref{eq:method-chunk-rep}) and
discards the chunk states, since the selected chunks are re-encoded during
generation. It therefore never runs layers $\ell_c+1,\dots,L$ on the
context, saving $2\Theta\lambda$ per context token.

\item[(b)] \emph{Attention to unselected chunks (UNREAL saving).} Each part counts
causal query-key pairs, at $4L_a d_a$ FLOPs per pair.
\begin{itemize}
\item \emph{Prefill.} The full-context prefill computes all $N(N+1)/2$ causal
pairs among the context tokens. UNREAL's independent chunk encoding computes
only the $M\,T(T+1)/2 = N(T+1)/2$ intra-chunk pairs. The remaining
$N(N-T)/2$ cross-chunk pairs give the first term.
\item \emph{Query.} In full context, each of the $T_q$ query tokens attends
to all $N$ context tokens. In UNREAL it attends only to the $nT$ selected
tokens, a difference of $T_q(N-nT)$ pairs.
\item \emph{Decode.} Likewise, each of the $T_g$ decoded tokens attends to
$N - nT$ fewer keys, a difference of $T_g(N-nT)$ pairs.
\end{itemize}

\item[(c)] \emph{Re-encoding the selected chunks (UNREAL extra cost).} UNREAL's
generation prefill passes the $nT$ selected tokens through all layers again
($2\Theta\,nT$) and computes their $nT(nT+1)/2$ causal self-attention
pairs. The query's attention to these tokens is already counted in (b).

\item[(d)] \emph{Retrieval pass (UNREAL extra cost).} The forward pass on
$x_{\text{ret}}$ has no counterpart in full-context inference. Its tokens
cost $2\Theta\,|x_{\text{ret}}|$ in weights plus
$|x_{\text{ret}}|(|x_{\text{ret}}|+1)/2$ attention pairs.

\item[(e)] \emph{Intra-chunk attention above $\ell_c$ (UNREAL saving).}
This restores the full-context-only intra-chunk work in the
$L_a-L_{a,\leq c}$ global-attention layers above the readout.
\end{itemize}

The $+1$ parts of (c)-(d) are diagonal-pair corrections, only
$1/\kappa$ of the corresponding weight costs, where
$\kappa\equiv\Theta/(L_a d_a)$. Term (e) is a second correction, across the three backbones and $n\in\{5,10,20\}$, omitting it raises the
roots by at most $0.21\%$, so we omit it below.

\subsubsection{Simplifying assumptions}
We now reduce \eqref{eq:flops-delta} under two assumptions:
\begin{enumerate}
\item[(S1)] \emph{Short queries and generations:} $T_q = T_g = T$.
\item[(S2)] \emph{The context is much longer than the selected evidence:}
$N \gg nT$. Since $n\geq1$, this also implies $N\gg T$. For fixed, moderate
$n$, $|C_0(x)|$, and $R/T$, it further gives
$N\gg|x_{\text{ret}}|$, since
$|x_{\text{ret}}|=(|C_0(x)|+1)T+R=O(nT)$.
\end{enumerate}

\begin{equation}
\Delta \approx
\underbrace{2\Theta\lambda N}_{\text{deeper layers}}
+ \underbrace{2L_a d_a N^2}_{\text{cross-chunk attention}}
- \underbrace{2\Theta\bigl(nT + |x_{\text{ret}}|\bigr)}_{\text{re-encoding and retrieval pass}}.
\label{eq:flops-delta-simp}
\end{equation}
The neglected terms have relative size of order $T/N$ and $(nT/N)^2$, which
vanish in the long-context limit. The diagonal corrections are also smaller
than their weight terms by $L_a d_a/\Theta=1/\kappa$; additionally,
$\kappa\gg1$ for all evaluated backbones. Because these dropped terms have mixed signs,
Eq.~\eqref{eq:flops-delta-simp} is an approximation, not a bound.

\subsubsection{Break-even context length}
Dividing Eq.~\eqref{eq:flops-delta-simp} by $2L_a d_a$, UNREAL requires fewer FLOPs
($\Delta > 0$) if and only if
\[
N^2 + \lambda\kappa N - \kappa\bigl(nT + |x_{\text{ret}}|\bigr) > 0,
\]
that is, for $N > N^*$ with
\begin{equation}
N^* = \tfrac12\Bigl(-\lambda\kappa + \sqrt{\lambda^2\kappa^2 + 4\kappa\bigl(nT + |x_{\text{ret}}|\bigr)}\Bigr).
\label{eq:flops-nstar}
\end{equation}

\paragraph{Last-layer readout.} For $\ell_c = L$ (i.e.\ $\lambda = 0$),
Eq.~\eqref{eq:flops-nstar} reduces to
\begin{equation}
N^* = \sqrt{\kappa\bigl(nT + |x_{\text{ret}}|\bigr)}.
\label{eq:flops-nstar-last}
\end{equation}
This is the length at which the avoided cross-chunk attention, about
$2L_a d_a N^2$, equals the weight cost of the tokens UNREAL processes
additionally. Any $\lambda > 0$ adds the linear saving (a) and lowers $N^*$
below this value.

\paragraph{Self-consistency.} For $\lambda=0$, let
$Q=nT+|x_{\text{ret}}|$. The condition $Q\ll\kappa$ gives
$N^*/Q=\sqrt{\kappa/Q}\gg1$, so the predicted root satisfies (S2). This
generally holds for large LLMs with modest selected evidence.

\paragraph{Longer generations.} When $T_g\gg T$, term (b) adds
$4L_a d_aT_g(N-nT)$ FLOPs of savings, lowering the break-even length toward
$nT$ from above.

\subsection{\texorpdfstring{$N^*$}{N*} for the evaluated backbones}
We evaluate \eqref{eq:flops-nstar} for the three 30B-scale backbones,
using their released configurations and the per-backbone $\ell_c$ of
Appendix~\ref{app:training-configuration}.

For all three backbones we use $T = 141$ (the median chunk length in
Nemotron tokens), $|C_0(x)| = 5$, $R = 64$, and $T_q = T_g = T$, so
$|x_{\text{ret}}| = 910$. Table~\ref{tab:flops-nstar} reports the results.

\begin{table}[h]
\centering
\small
\caption{Approximate break-even context length $N^*$ (tokens;
\eqref{eq:flops-nstar}), with $T = 141$, $T_q = T_g = T$, and
$|x_{\text{ret}}| = 910$. $\Theta$ counts active non-embedding matrix
parameters.}
\label{tab:flops-nstar}
\setlength{\tabcolsep}{4pt}
\begin{tabular}{lccccccc}
\toprule
 & & & & & & \multicolumn{2}{c}{$N^*$ (tokens)} \\
\cmidrule(lr){7-8}
Backbone & $\Theta$ & $L_a$ & $\kappa$ & $\ell_c/L$ & $\lambda$ & $n = 5$ & $n = 10$ \\
\midrule
Muse-Glimmer-30B & 25.2B & 13 & $4.73\times10^{5}$ & 47/52 & 0.096 & 13{,}049 & 17{,}437 \\
Nemotron-3.5-Lightning & \phantom{0}2.87B & \phantom{0}6 & $1.17\times10^{5}$ & 42/52 & 0.201 & \phantom{0}6{,}321 & \phantom{0}8{,}471 \\
Qwen3.5-35B-A3B & \phantom{0}2.44B & 10 & $5.95\times10^{4}$ & 27/40 & 0.323 & \phantom{0}4{,}117 & \phantom{0}5{,}569 \\
\bottomrule
\end{tabular}
\end{table}

Across these settings, \eqref{eq:flops-nstar} is within $0.55\%$ of the
numerical root of the full \eqref{eq:flops-delta}.

\section{Efficiency benchmark: methodology and full results}
\label{app:efficiency}

We report measured time-to-first-token (TTFT) and
throughput for UNREAL's chunk-then-select pipeline against full-context
inference. This section gives the full methodology, the FLOPs model, and
results for all four backbones from \cref{sec:retrieval}.

\textbf{Setup.} We benchmark performance only, not accuracy: all models are
randomly initialized (\texttt{load\_format="dummy"}) and all inputs are
random token ids (\texttt{skip\_tokenizer\_init}), so no checkpoints or
tokenizers are required. We serve each backbone with
vLLM~\citep{kwon2023efficient} rather than a plain \texttt{transformers} forward
pass, so results reflect PagedAttention, FlashAttention-3, and continuous
batching. We use the chunking regime evaluated throughout \cref{sec:longctx}
($T{=}128$, $n{=}10$) and sweep $L$ over the same ten values as
\cref{fig:teaser-figure} (8K--100M tokens).

All measurements use bf16 weights and activations on a single NVIDIA H100
80GB GPU, except Qwen3.5-35B-A3B's KV cache, stored in fp8 since its
$\approx$64GB of resident weights (256 routed experts) leave too little
bf16 headroom for a 256K-token cache. With this, all four baselines are
measured directly up to 256K tokens. vLLM dispatches each
backbone's optimized fused kernels (FlashInfer's Gated-DeltaNet for linear
attention, native Mamba2/MoE mixers for Nemotron-3.5-Lightning)

\textbf{FLOPs model.} We count prefill multiply-add FLOPs as 2 FLOPs/MAC,
walking each backbone's \emph{real} per-layer sequence rather than a single
network-wide ratio. Every layer contributes a linear ($O(L)$) projection
cost -- QKVO for full-/sliding-attention, or the real linear-attention /
Mamba2 projection sizes otherwise
plus its feed-forward cost (dense or MoE, charging only the active and any
always-on shared experts, with 2 or 3 weight matrices matching each
backbone's real gated/non-gated activation). Full-attention layers
additionally cost $C_{\mathrm{attn}}(L){=}2L^2 d_{\mathrm{attn}}$
($d_{\mathrm{attn}}{=}n_{\mathrm{heads}} d_{\mathrm{head}}$); sliding-attention
layers cost $2 L \min(L,W) d_{\mathrm{attn}}$ for window $W$ -- linear in
$L$ once $L{>}W$, the same logic that makes UNREAL's own $T$-chunked
encoding linear; linear-attention and Mamba2 layers add no quadratic term.
The LM head is counted only on passes that emit logits. Summed over the
real layer sequence, full-context inference costs
$\mathrm{FLOPs}_{\mathrm{base}}(L) = O(L) + O(L^2)$, while UNREAL costs
$\mathrm{FLOPs}_{\mathrm{UNREAL}}(L) =
\underbrace{O(L) + O(L\, T)}_{\text{encode, through }\ell_c}
+\ \underbrace{O(M)}_{\text{score}}
+\ \underbrace{O(1)}_{\text{query}}
+\ \underbrace{O(1)}_{\text{generate}}$ ($M{=}L/T$): the quadratic term
shrinks to $L\cdot L_C$, and generation always reads exactly $n T{=}1{,}280$
tokens regardless of $L$.

\begin{figure}[t]
  \centering
  \includegraphics[width=\linewidth]{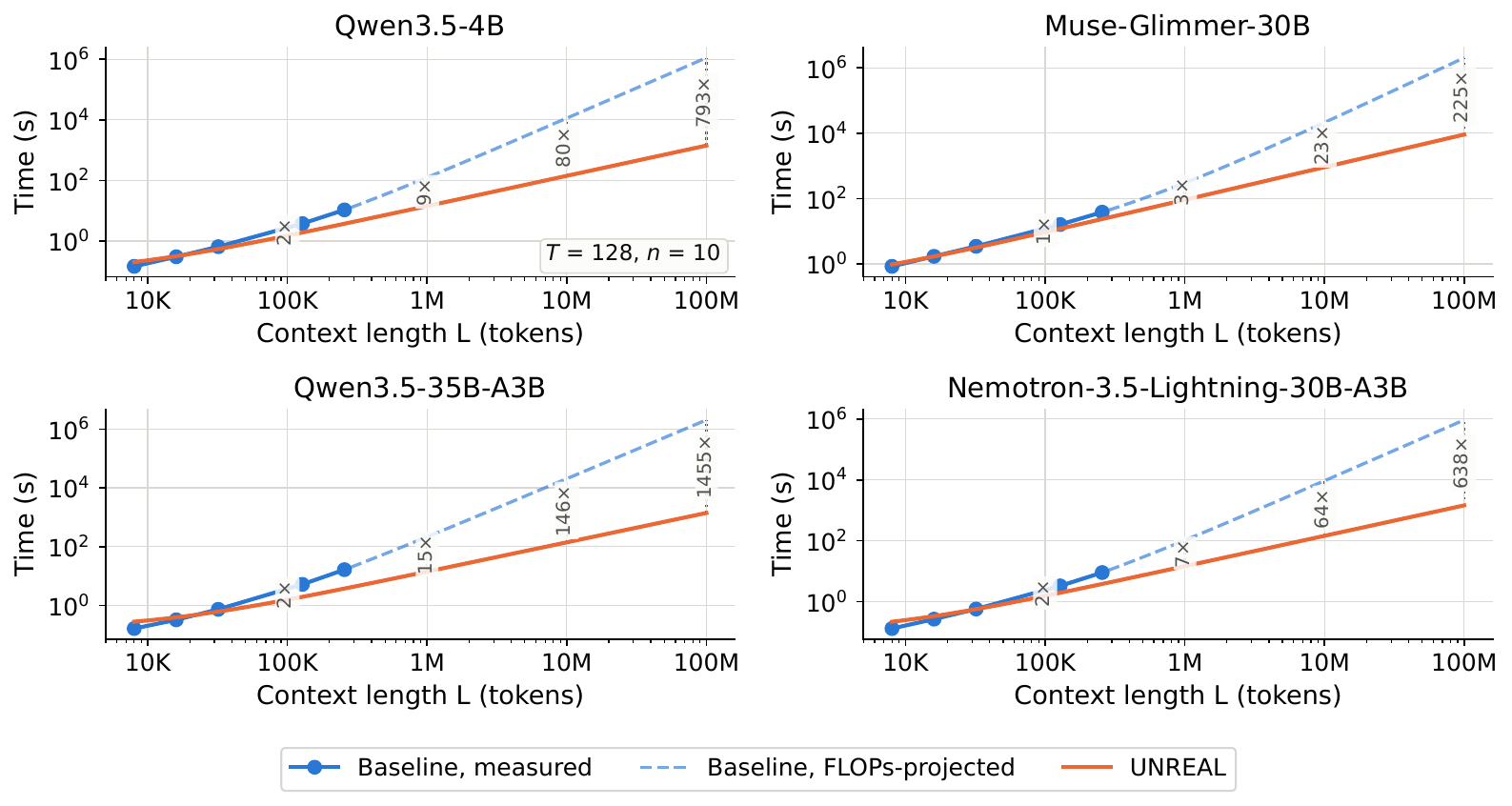}
  \caption{Measured time-to-first-token, baseline vs.\ UNREAL, for all four
  backbones (random weights/inputs, served with vLLM on an NVIDIA H100,
  bf16). Solid markers are directly measured; the dashed baseline segment is
  a FLOPs-scaled projection past the point where a single full-context
  forward pass exceeds our measurement budget or the available KV cache.
  Dotted vertical connectors mark the speedup (baseline time / UNREAL time)
  at a few representative context lengths, read directly off the time gap
  between the two curves.}
  \label{fig:ttft-all-backbones}
\end{figure}

\begin{figure}[t]
  \centering
  \includegraphics[width=\linewidth]{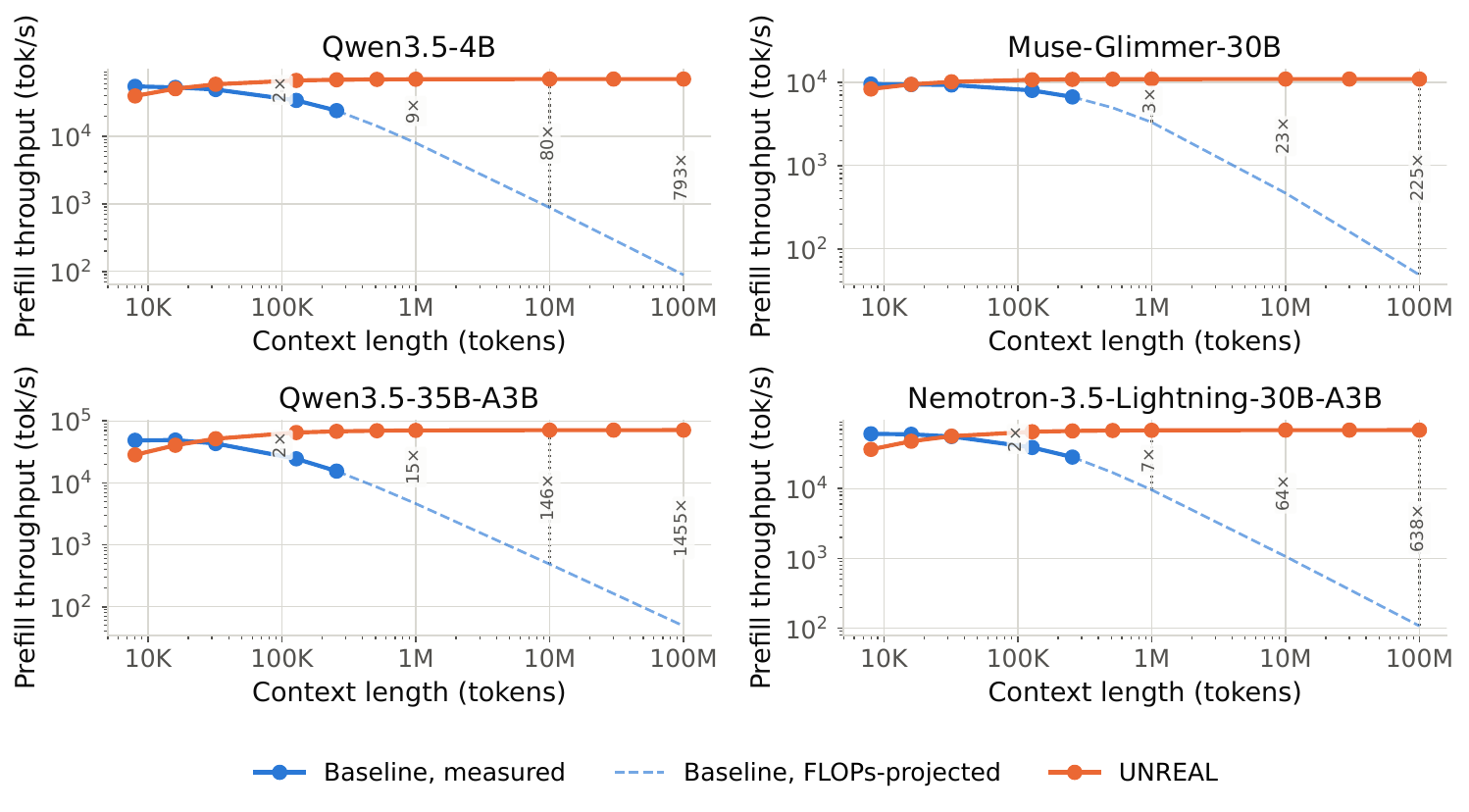}
  \caption{Measured prefill throughput (tokens/second), baseline vs.\
  UNREAL, for all four backbones. Same data and conventions as
  \cref{fig:ttft-all-backbones}.}
  \label{fig:throughput-all-backbones}
\end{figure}

\textbf{Measured wall-clock results.} \cref{fig:ttft-all-backbones,fig:throughput-all-backbones}
report measured TTFT and throughput for all four backbones. The baseline
curve is shown only up to the context length at which a single forward pass
exceeds our 75-second measurement budget (solid markers), matching the
convention of \cref{fig:teaser-figure}; we extend it with a dashed,
FLOPs-scaled projection ($t_{\mathrm{proj}}(L) = t_{\mathrm{measured}}(L_{\mathrm{cutoff}})
\cdot \mathrm{FLOPs}_{\mathrm{base}}(L)/\mathrm{FLOPs}_{\mathrm{base}}(L_{\mathrm{cutoff}})$)
for visual continuity across the same 8K--100M range. UNREAL's chunk-encoding
time is computed from its measured, steady-state per-chunk throughput.
Chunk encoding is parallel across the $M{=}L/T$ chunks, so
total encode time scales linearly with $M$ plus the directly measured,
constant-size query and generation passes. This measured
$T_{\mathrm{UNREAL}}(L)$ omits two components, both excluded because they
are negligible or out of scope rather than favorable to UNREAL: (1) the MaxSim
scoring step ($O(M)$ in the FLOPs model above) is not separately
benchmarked, as it is a lightweight vector similarity computation external
to the LLM forward pass that accounts for under $0.01\%$ of UNREAL's total
FLOPs at every backbone and length we tested; and (2) any orchestration
overhead between the encode and generate stages (selecting the top-$k$
chunks, assembling the generation prompt), which a production serving
stack would incur but which our per-stage vLLM benchmarking (each stage
measured on its own, isolated engine) does not capture.
The measured curves corroborate
the FLOPs estimate: UNREAL's TTFT grows roughly linearly with $L$ while the
baseline's grows quadratically, so the baseline becomes both slower and,
eventually, infeasible to run at all, in the same context-length regime
where \cref{fig:teaser-figure} shows its accuracy collapsing.

